%% file: main_arxiv.tex
\documentclass{article}

\PassOptionsToPackage{numbers, compress}{natbib}
\usepackage[preprint]{neurips_2026}

\usepackage[utf8]{inputenc} 
\usepackage[T1]{fontenc}    
\usepackage{hyperref}       
\usepackage{url}            
\usepackage{booktabs}       
\usepackage{multirow}       
\usepackage{amsfonts}       
\usepackage{nicefrac}       
\usepackage{microtype}      
\usepackage[dvipsnames]{xcolor}         
\usepackage{fontspec}
\usepackage{graphicx}
\usepackage[most]{tcolorbox}
\usepackage{enumitem}
\hypersetup{colorlinks=true, urlcolor=blue, linkcolor=black, citecolor=black}
\usepackage{float}

\AtBeginDocument{\newgeometry{hmargin=1.25in, bottom=1.25in, top=1in, footskip=30pt}}

\newfontfamily\myfont{Silkscreen-Bold.ttf}

\newif\ifarxiv
\arxivtrue

\title{{\myfont \LARGE Ask-E:} An Environment for\\Calibrated Question Generation}

\author{%
  Sarah Pratt\textsuperscript{1}\thanks{Correspondence to \texttt{spratt3@uw.edu}}
  \And
    Jae Sung Park\textsuperscript{2}
  \And
    Scott Geng\textsuperscript{1}
  \And
    Ali Farhadi\textsuperscript{1}
  \AND
        \normalfont\textsuperscript{1}University of Washington \quad \textsuperscript{2}Allen Institute for AI
}

\begin{document}

\maketitle

\begin{abstract}
\input{sections/abstract}
\end{abstract}

\begin{center}
\textit{``Computers are useless. They can only give you answers.''} \\
--- Pablo Picasso
\end{center}
\vspace{0.3cm}

\input{sections/intro}

\input{sections/related_work}

\input{sections/methods}
\input{sections/benchmark}
\input{sections/training}
\input{sections/conclusion}

\renewcommand{\acksection}{\section*{Acknowledgements}}
\begin{ack}
We thank Vivek Ramanujan, Aditya Kusupati, Hamish Ivison, Nabil Omi, Reza Salehi, and Matthew Wallingford for their helpful discussions, feedback, and advice throughout this project. We also thank the Hyak computing cluster team at the University of Washington for their support with the computing infrastructure that made this work possible. This work was supported in part by a Gemini Academic Program Award from Google.
\end{ack}

\newpage
\bibliographystyle{plainnat}
\bibliography{main}


\appendix

\newpage
\input{supp/limits}
\input{supp/prompts}
\input{supp/full_boundary_models}
\input{supp/rollout_ex}

\input{supp/training_deats}
\input{supp/bench_deats}
\input{supp/additional_analyses}
\input{supp/downstream_stuff}
\input{supp/full_table}


\end{document}

%% file: sections/abstract.tex
Today, we improve models by training and evaluating them on problems at the frontier of their abilities. Creating such problems is itself a demanding task, requiring the ability to probe model limits and generalize beyond existing question distributions. It also means placing problems at a precise difficulty level, which requires understanding what it takes to solve them. In short, generating problems calibrated to a model's current frontier demands capability beyond it, an increasingly burdensome constraint as models improve. Our key insight is that we can leverage this constraint to our advantage: a model that can generate problems consistently calibrated to a given frontier must possess capability beyond it. Accordingly, we present Ask-E, an environment that benchmarks and trains models on their ability to write questions at a given skill level, rather than answer them. Concretely, we define target skill levels as ranges bounded by the capabilities of two existing language models. A generated question is successfully calibrated if exactly one of the two models can solve it, placing it precisely within the target range and differentiating the capabilities of these models. Ask-E serves both as a benchmark and a training environment, where models generate problems calibrated to a variety of skill levels. We find that even frontier models achieve below 50\% calibration on the benchmark, leaving significant headroom to measure future progress. We also show that training on this environment leads to improvements across a number of downstream math benchmarks even with no new math data, no interaction with stronger models, and no correctness-based reward.
\begingroup
\renewcommand\thefootnote{}\footnotetext{Code and released rollouts: \url{https://github.com/sarahpratt/aske}}%
\addtocounter{footnote}{-1}%
\endgroup

%% file: sections/intro.tex
\section{Introduction}

\begin{figure*}[ht]
  \centering
  \includegraphics[width=\textwidth]{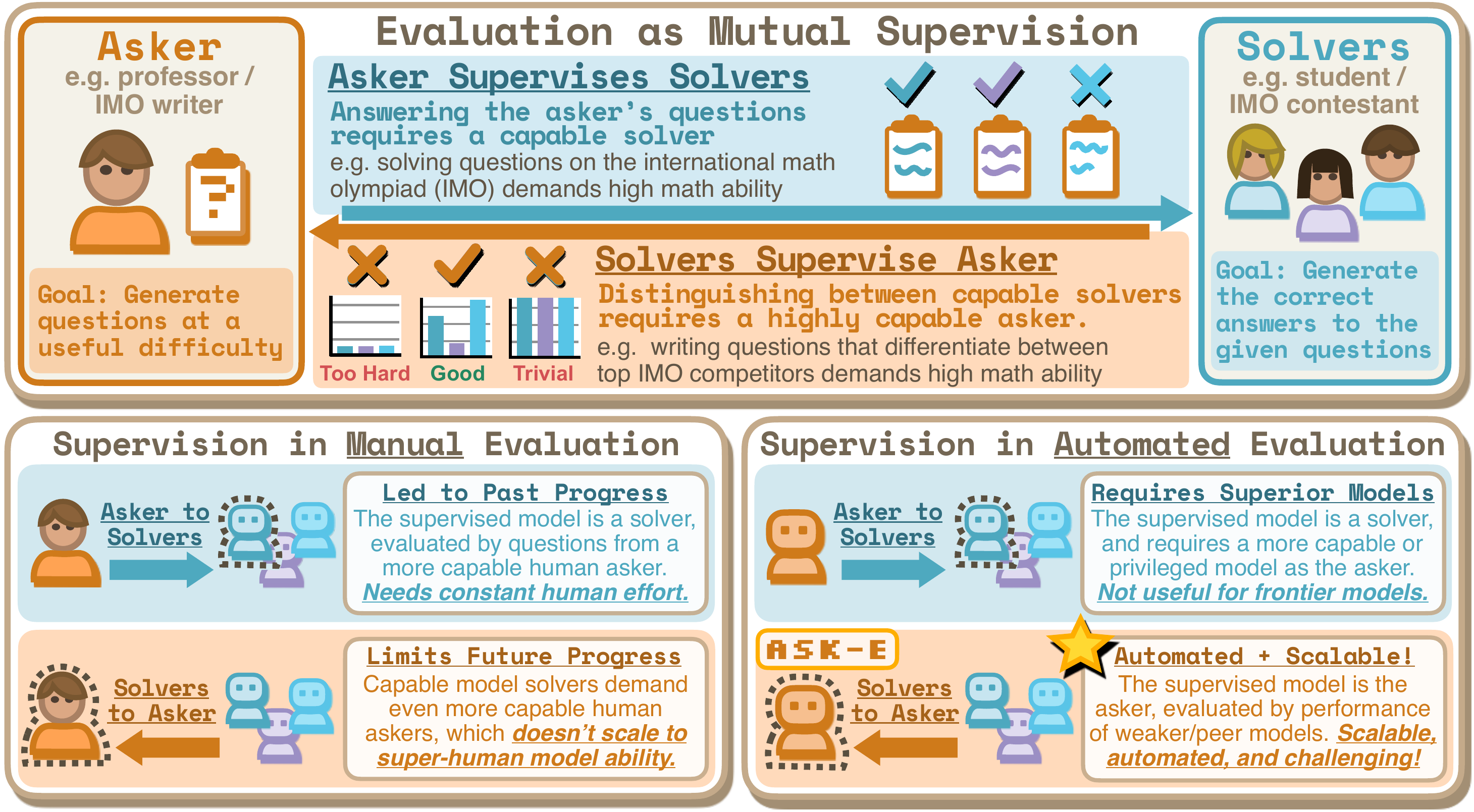}
   \caption{\textbf{Evaluation as mutual supervision.} Answering hard questions clearly demands skill, but so does writing them at the right difficulty. Solving an IMO problem requires deep mathematical ability, and so does writing a problem that differentiates between top IMO contestants. This mutual demand presents a problem for standard evaluation, where improving models require increasingly capable question askers to produce questions at the right difficulty for evaluation and training. Ask-E takes advantage of this dynamic by instead testing models as the question asker, a role supervised by the performance of weaker or peer solvers.
}
   \label{fig:teaser}
\end{figure*}
 
The dominant force behind language model improvement in domains like mathematics has been measuring their ability to solve existing problems. The resulting performance is then used either as a training signal or as a benchmark to measure progress. However, this approach depends on a supply of problems at the right difficulty level. Problems that are too easy for today's frontier, such as grade school math, do nothing to differentiate between state-of-the-art models. Problems that are too hard, such as solving the Riemann hypothesis or correctly guessing a number between one and a million, are equally uninformative, as all models will fail regardless of their relative skill.
 
Obtaining questions at the right difficulty level is not only crucial, it is very difficult. The best models currently perform on par with world-class high school students, meaning that researchers must find math competition champions, university professors, and research mathematicians to continue to challenge them. Problem authors must be able to effectively probe models to understand their abilities, requiring knowledge of what is useful to ask and how to interpret the responses. They must have mathematical understanding beyond that of the models to identify their current shortcomings. They must possess creativity to write questions sufficiently different from existing distributions to identify true generalizable skills. Finally, problem authors must understand the skills and knowledge needed at each step from the problem statement to the solution in order to effectively target the capabilities they hope to measure or elicit.
 
In this work, we posit that the challenging nature of producing well-calibrated questions may itself provide a path forward. As previously noted, writing a question that distinguishes models of different abilities requires all the skills needed to answer that question, and more. Additionally, a question author's success is judged not by a more capable supervisor, but by the evaluated models themselves. If the evaluated models differ in their ability to solve the question, the author has succeeded. If all models answer correctly or all answer incorrectly, the author has failed. The question writer, performing the harder task, is indirectly graded by the solvers, performing the easier one (Figure \ref{fig:teaser}).
 
\begin{figure*}[t]
  \centering
  \includegraphics[width=\textwidth]{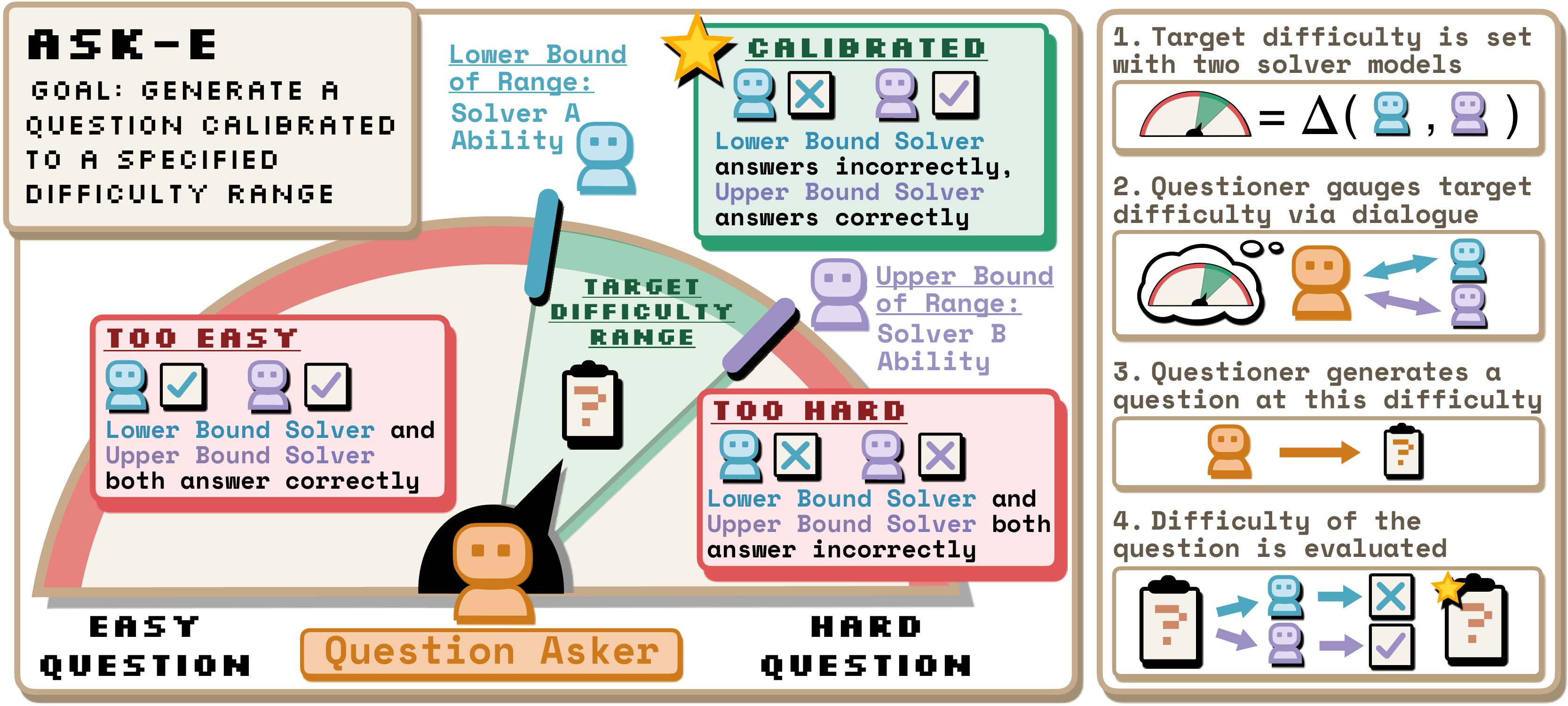}
   \caption{\textbf{Overview of the Ask-E environment.} The question asker must generate a question calibrated to a target difficulty range, whose lower and upper bounds correspond to the ability levels of two solver models. A question is too easy if both boundary solvers answer correctly, too hard if both answer incorrectly, and calibrated if exactly one succeeds, placing its difficulty inside the target range. (Right) The asker gauges the target range through dialogue with the two solvers before generating a final question at the target difficulty.
}
   \label{fig:aske}
\end{figure*}

Motivated by these observations, we present Ask-E, an environment 
where the goal is to do exactly what researchers have themselves been doing for decades: generate questions that will differentiate the capabilities of existing models. In Ask-E, a question asker model interacts with various pairs of solver models sampled across a broad range of capabilities. The asker first probes the models in the pair to identify their capability difference, and then produces a final question that targets this span between the two models' skill levels. The asker model is successful in an interaction with a given pair if it produces a question that one model can answer correctly and one cannot (Figure \ref{fig:aske}).

Ask-E offers three advantages over the standard question-answering approach. First, it provides a more demanding evaluation paradigm. As models improve, the standard response is to find harder questions for them to answer. We instead change the task itself, asking models to do something harder than answering questions. Second, Ask-E naturally scales with the field. While improved model capabilities saturate most benchmarks, in Ask-E better models just represent more advanced skill levels to target, meaning the task naturally gets harder as models get better. Finally, it offers a path forward as models surpass human ability. Standard approaches require a supervisor more capable than the evaluated model, a constraint that becomes unsustainable as models improve. Ask-E inverts this dynamic: the supervisory signal can come from models of equal or lesser ability, whose answer variance reveals whether the evaluated model has successfully calibrated its question.
 
Our contributions are as follows:
\begin{itemize}[leftmargin=*]
  \item \textbf{The Ask-E environment.} We present an environment for measuring and developing a model's ability to generate calibrated questions. Concretely, we define target difficulty ranges bounded by the capabilities of two existing LMs. An asker model must interact with these boundary solvers to identify a gap in their abilities and then calibrate a question to that gap. The asker succeeds if one model is able to solve the question and the other is not, differentiating their skill level.
  \item \textbf{Benchmarking on Ask-E.} We evaluate a range of models on calibrated question generation and find that performance correlates with general model ability. State-of-the-art models achieve only 44.9\% success rate, demonstrating substantial room for improvement. Furthermore, the benchmark is designed to evolve: as models saturate the current boundary set, stronger boundary solvers can be introduced to raise the difficulty ceiling.
  \item \textbf{RL training on Ask-E.} We show that skills learned by asking questions transfer to answering them. We train a model solely on calibrated question generation, with no external math data, no interaction with stronger models, and no reward for the correctness of its own answers. Despite this, the trained model improves at question answering on challenging math benchmarks including AIME, HMMT, and IMO-AnswerBench, supporting the view that calibrating questions demands the skills needed to solve them.
\end{itemize}

%% file: sections/related_work.tex
\section{Related Work}

\paragraph{LMs for Question Generation.}
Prior work has explored training models to generate questions for a variety of practical applications. In the area of education, for example, a number of works \citep{wang2018qg, srivastava2021question, al2024analysis, elkins2023useful, zhuge2025twinstar} have explored the ability of LMs to generate educational questions to aid in student learning. Effective question generation has also been studied as a tool to resolve ambiguity in human preferences and instructions in human-LM conversations \citep{mazzaccara2024learning, wang2025learning, li2025questbench, pedrozo2026reasoning}. These works point to the potential of LMs to not only provide information, but to learn to ask useful questions. However, in this work, we take this one step further. We view question asking not only as a means to an end, but as a skill which can measure complex reasoning abilities within models as well as a training task to reinforce these abilities.

\paragraph{LMs for Automated Benchmarks.}
LMs have also been increasingly used to generate evaluation benchmarks for other LMs. As models grow more capable, hand-crafted benchmarks require greater expertise to construct \citep{garre2026riemann, luong2025towards} and saturate more quickly \citep{kiela2021dynabench}. One response to this problem has been to develop algorithmic methods that systematically generate harder questions through templates and composition \citep{zhu2023dyval, mirzadeh2024gsm, zhou2025gsm, sun2025omega, srivastavabeyondbench}. These approaches can be more robust than static benchmarks, but remain limited in the scope of what they can evaluate. Other works use LMs themselves to generate challenging benchmarks. AutoBencher \citep{li2024autobencher} iteratively proposes and refines questions to satisfy properties like difficulty or novelty. MATH² \citep{shah2024ai} generates novel questions by combining pairs of skills extracted from existing benchmarks. CHASE \citep{patel2025get} composes hard problems from simpler, verifiable components rather than generating them directly. In these frameworks, question generation is a means of constructing the benchmark, and a model is evaluated by whether its answers match the generating LM's answers. 

We instead treat question generation as the capability being evaluated, assessing whether a model's generated questions match a target LM's ability level. This avoids two drawbacks of standard automated benchmarks. First, in standard settings, the supervising LM must be more capable than the evaluated model, as it must generate the question and answers while the evaluated LM just generates the answer. This means automated benchmarks can only evaluate models below the capability of the LM that generated them, limiting the usefulness for benchmarking frontier models. In Ask-E, this relationship is inverted. The evaluated model performs the harder task of question generation, while the supervising models perform the easier task of answering. This allows the benchmark to be fully automated without necessarily being solvable by an existing LM. Second, standard automated benchmarks rely on LM-generated answers as ground truth, which can be noisy or incorrect \citep{feuer2025judgment}. In Ask-E, ground truth is not a generated answer but the boundary solvers' own performance, which is an observable property rather than a generated artifact. As models continue to improve, automated evaluation will become increasingly necessary, and we see our approach as a natural step in this direction.

\paragraph{LMs for Answer Evaluation.}

Even with human-authored benchmarks, it is increasingly common for responses to be graded by LMs \citep{chernyshev2024u, luong2025towards}, and this has been explored during training as well \citep{yuan2024self, bai2022constitutional}. In these settings, the grading LM is a tool for evaluation, not itself being evaluated. However, several recent works explicitly evaluate an LM's ability to perform tasks beyond answering questions. IMO-GradingBench \citep{luong2025towards} evaluates whether LMs can accurately score mathematical proofs, and RewardBench \citep{lambert2025rewardbench} evaluates reward models on their ability to rank responses. These works highlight that tasks beyond answering questions also require genuine skill, and a model's ability to perform them can be just as revealing of its underlying capabilities. The Delta Learning Hypothesis \citep{geng2025delta} goes further, showing that training a model to identify the better of two weak model responses improves the model's own capabilities in the underlying domain, demonstrating that performing evaluation can itself be a source of learning. We view our work as an extension of this direction, where rather than ranking model capabilities, our task requires targeting them precisely enough to generate a question that falls within a specific ability gap.

\paragraph{LMs for Play via Dialogue.}
A recent related line of research explores learning through competitive interaction between models. DTE \citep{srivastava2025debate} trains a model on multi-agent debate traces. SPIN \citep{chen2024self} uses a generator-discriminator framework where one copy attempts to produce text indistinguishable from human writing while another tries to tell them apart. CodeClash \citep{yang2025codeclash} has agents iteratively edit code based on how it performs against a competitor. SPIRAL \citep{liu2025spiral} has agents compete in zero-sum games like poker and tic-tac-toe, finding that this improves general reasoning abilities.

Most closely related to our work is a series of recent papers on self-play for calibrated question generation \citep{liu2025spice, yang2025spell, chen2025self, kuba2025language, zhao2025absolute, huang2025r}. In these frameworks, a model acts as both a challenger, generating questions, and a solver, answering them. The challenger is rewarded for generating questions at the edge of the solver's ability, and the solver is rewarded for answering correctly. These works differ from each other primarily in their chosen domain and in how they obtain ground truth for verification: grounding question-answer pairs in an external document corpus \citep{yang2025spell, liu2025spice}, generating and executing code \citep{zhao2025absolute}, using majority voting as a proxy for correctness \citep{huang2025r, chen2025self}, or relying on an external reward model \citep{kuba2025language}. All demonstrate that models can set meaningful challenges for each other and improve by solving them, without relying on human-authored questions.

\begin{figure*}[t]
  \centering
  \includegraphics[width=\textwidth]{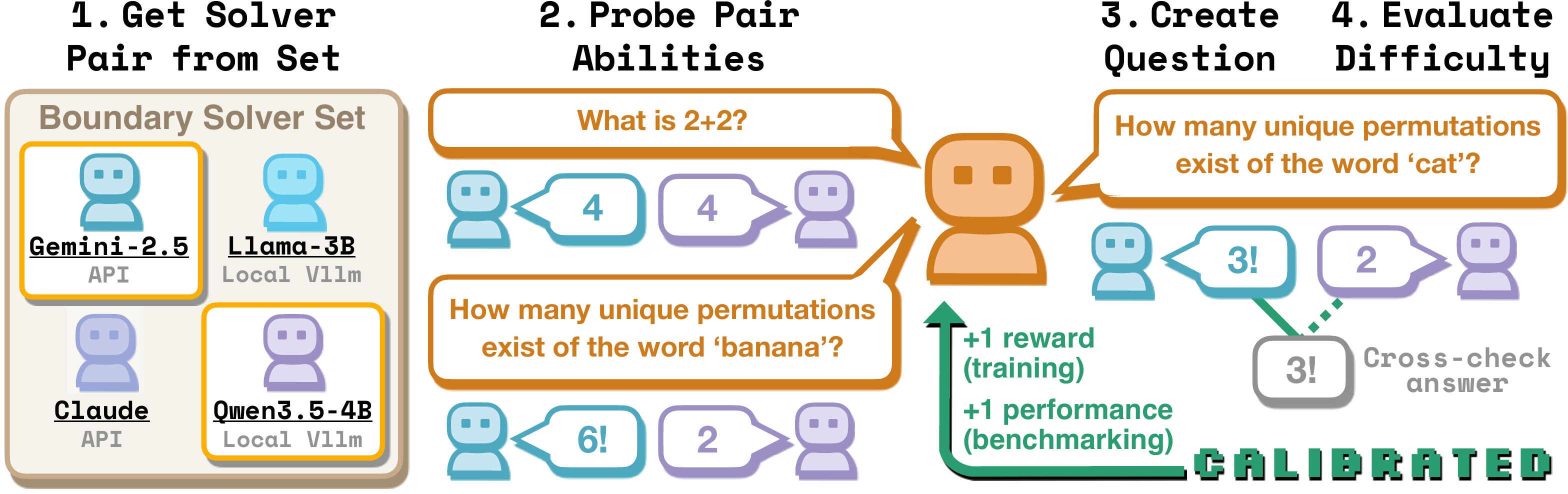}
   \caption{\textbf{A detailed walkthrough of the Ask-E environment for a single session.} (1) A boundary pair is selected from the boundary solver set, which mixes API-based and locally served models. (2) The asker probes the pair with questions and observes their responses to gauge the gap in their abilities. (3) The asker generates a final question designed to fall within this gap. (4) Both boundary solvers and the cross-check solver answer the question; the session is scored as calibrated if exactly one boundary solver matches the cross-check answer, corresponding to a reward during training or a success during benchmarking.}
   \label{fig:env}
\end{figure*}

However, in all of these works, question generation is a means to an end, not a measured capability. The challenger co-evolves with a single solver, so it need only track one moving target, and its questioning ability is never itself scored. In contrast, Ask-E fixes a diverse set of boundary solver pairs spanning a wide range of abilities, and measures the asker on its calibration to each. Because the targets are fixed, calibration rates are comparable within a training run and across model releases. Moreover, the large variety of targets means the task measures general question-generation ability rather than usefulness to any single model. The works discussed above point to two observations. Posing well-calibrated questions is challenging, potentially requiring skills that include and exceed those needed to answer them. Meanwhile, as models improve, automated evaluation becomes increasingly essential. Ask-E combines the two, treating this challenging task as the basis for a fully automated training environment and benchmark.

%% file: sections/methods.tex
\section{The Ask-E Environment}

Ask-E is an environment where models are presented with a target difficulty level, defined by the difference in ability levels of two existing LMs. A model succeeds by accurately gauging this target and generating a question at the specified difficulty. The environment consists of a question asker model, a set of boundary solvers, a cross-check solver, and a defined conversation pipeline. Each session consists of four steps, visualized in Figure \ref{fig:env}, and results in a binary outcome: successful calibration or failure.

In the first step, two boundary solvers are selected from the boundary set. Because each boundary solver is a distinct LM, the two models will generally differ in capability. The gap between their ability levels defines the target question difficulty for the session. The boundary set consists of $n$ LMs, so the environment supports $n \choose 2$ distinct difficulty ranges across sessions. A natural question is why the target must be defined by the ability of two models rather than one. A single solver provides only a one-sided constraint. If the goal is merely to write a question the solver answers correctly, an arbitrarily easy question suffices, and if the goal is a question it answers incorrectly, an impossibly hard one does. Neither demands skill from the asker. Knowing that a problem is too hard for a given solver does not require the ability to solve it (we know the Riemann hypothesis exceeds current human ability without knowing what its proof demands). Pinning a problem's difficulty from both sides is different. To place a question above one solver's ability and below the other's, the asker must gauge how hard the question actually is, which requires understanding the path to its solution. Two boundary solvers thus convert an open-ended constraint into a bounded target that cannot be satisfied trivially.

In the second step, the asker must assess the target difficulty level by probing the capabilities of the two boundary solvers. It does this through $k$ probing rounds, in which the asker poses a question to both boundary solvers and receives their responses along with a short summary of each model's approach. Throughout the probing phase, the asker retains the full session history and knowledge of the overall task, while the boundary solvers are stateless and simply answer each question in isolation. 

In the third step, the asker uses the full probing context to generate a final question calibrated to discriminate between the two boundary solvers. The asker's role ends here; it does not observe the boundary solvers' responses to the final question.

In the final step, the question is sent to both boundary solvers and the cross-check solver, each of which answers it independently with no additional task context. To ensure the question has a single stable answer, the cross-check solver solves the question three times independently, and the session is gradable only if all three of its answers agree. The session is scored as calibrated if exactly one boundary solver matches the cross-check answer. It does not matter which of the two boundary solvers is the correct one (i.e. which one acts as the `upper' or `lower' bound). More on this design choice in Appendix~\ref{sec:weaker_wins}. This scoring mechanism has a useful implication for the cross-check solver. In standard automated benchmarks, the cross-check/answer-key model must be at least as capable as the model being evaluated, since both are performing the same task. In Ask-E, the asker and the cross-check solver perform fundamentally different tasks: the asker generates a calibrated question, while the cross-check solver only solves it. The cross-check model therefore only needs to be as capable as the solver models, and may be equal to or inferior to the asker model. It may even be the same model type as the asker, so long as it is a separate instance, ensuring the evaluated model cannot influence its own scoring.

We implement Ask-E using the Verifiers library \citep{brown_verifiers_2025}, which supports efficient multi-turn environments for LMs. We limit the scope of this work to math questions, though the environment could readily extend to other domains by changing the model prompts and answer equivalence libraries. Full prompts for the asker, boundary, and cross-check solvers can be found in Appendix \ref{sec:prompts}.

\begin{figure*}[t]
  \centering
  \includegraphics[width=\textwidth]{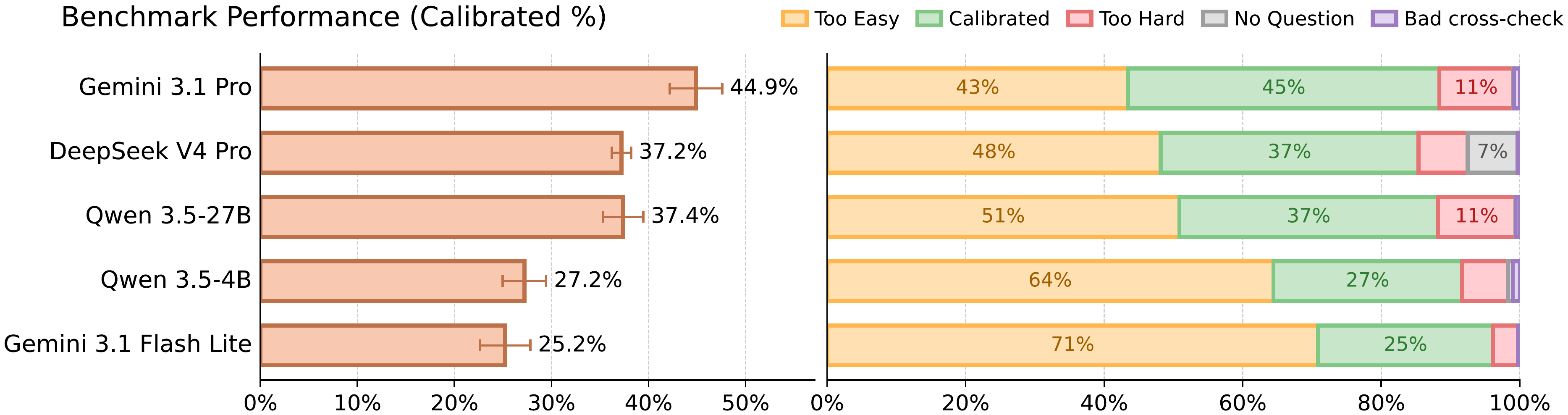}
   \caption{\textbf{Benchmark performance of five models on Ask-E.} (Left) Calibration rate across all sessions. Error bars correspond to 95\% confidence intervals computed across 10 independent runs of all 190 boundary pairs. (Right) Breakdown of generated question outcomes into too easy, calibrated, too hard, no question (no parseable final question), and bad cross-check (the three cross-check answers disagree) categories.
   }
   \label{fig:perf}
\end{figure*}

%% file: sections/benchmark.tex
\section{Benchmarking on Ask-E}
\label{sec:benchmark}

A primary use case of this environment is for benchmarking models. Calibration of ability level and generation of novel problems at a targeted level is a challenging task requiring skills including and beyond those required for standard question answering.

\subsection{Benchmark Implementation}
 
\paragraph{Boundary-solver sets.}
Our benchmark uses 20 boundary solvers: 17 open-weight checkpoints ranging from 1.5B to 27B parameters, plus 3 closed-API Gemini Flash models (Table~\ref{tab:boundary-sets}). This yields $\binom{20}{2} = 190$ unordered pairs. Each asker model interacts with every pair 10 times, for a total of $190 \times 10 = 1900$ sessions per evaluated asker. Boundary solvers are queried with $\texttt{temperature}=0.7$; all other sampling parameters are left at the backend's default (vLLM defaults for open-weight models, API defaults for the Gemini models). Where applicable we set thinking effort to \texttt{low} and disable thinking entirely on models that expose a switch (\texttt{enable\_thinking=false}) to encourage brevity. Open-weight models are served locally via vLLM \citep{kwon2023efficient} on 5 H200 GPUs (141 GB each); closed Gemini Flash models are queried through Google's OpenAI-compatible endpoint.
 
\paragraph{Dialogue specifications.}
Each session consists of 4 probing rounds followed by 1 final round. The asker is instructed to structure each reply into three sections, marked by \texttt{\#Reasoning\#}, \texttt{\#Draft\#}, and \texttt{\#Question\#} tags, where it thinks through its next steps, drafts candidate questions and solutions, and outputs a final question respectively. Only the question is forwarded to the boundary solvers. Boundary solvers respond with a solution and boxed answer, of which only a short summary is forwarded back to the asker. Full dialogue specifications including token budgets and parsing details are in Appendix~\ref{sec:more_bench}.
 
\paragraph{Scoring.}
We use Gemini~3.1~Pro \citep{gemini31pro2026} as the cross-check solver, which solves each final question three times independently. Sessions where the three cross-check answers disagree with each other are labeled \emph{bad cross-check} (under 1\% of sessions for every asker), filtering out questions without a stable, unambiguous answer. Final answers are extracted from boxed expressions and compared using the \texttt{math-verify} library \citep{kydlicek2025mathverify} with string-equality fallbacks, followed by an LLM equivalence check that catches formatting mismatches (e.g., ``1'' vs.\ ``$x=1$''). A session is labeled \emph{calibrated} when exactly one boundary solver matches all three cross-check answers, \emph{too\_easy} when both do, and \emph{too\_hard} when neither does. Full scoring details are in Appendix~\ref{sec:more_bench}, and Appendix~\ref{sec:opus_check} shows that results are robust to the choice of cross-check solver.

\begin{figure*}[t]
  \centering
  \includegraphics[width=\textwidth]{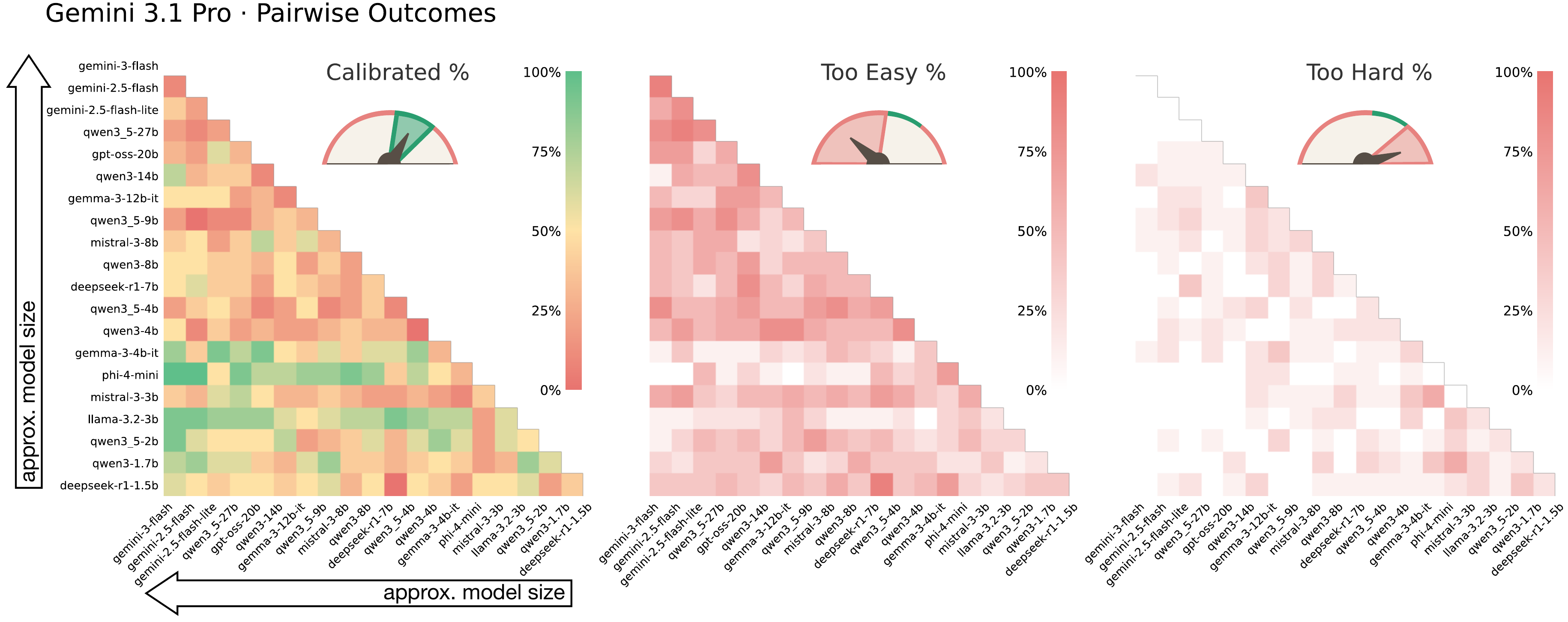}
   \caption{\textbf{Pairwise calibration outcomes for Gemini 3.1 Pro across all 190 boundary solver pairs.} Boundary solvers are ordered by approximate model size, increasing upward and leftward (arrows). The left heatmap shows calibration rate, which is highest when the pair consists of one small and one large model (bottom-left region), representing wide target ranges. Performance is moderate for pairs of small models (bottom-right region), which define narrow but easier targets, and lowest for pairs of large models (upper region), where the target is both narrow and requires generating highly challenging questions.
   }
   \label{fig:heatmap}
\end{figure*}

\subsection{Benchmark Results}

\paragraph{Model Performance.}
We evaluate 5 models on our proposed benchmark that span a wide range of abilities: Gemini 3.1 Pro \citep{gemini31pro2026}, Gemini 3.1 Flash Lite \citep{gemini31flashlite2026}, DeepSeek V4 Pro \citep{deepseekai2026deepseekv4}, Qwen3.5-27B \citep{qwen3.5} and Qwen3.5-4B \citep{qwen3.5}. Their performance is given in Figure~\ref{fig:perf}. Of the examined models, Gemini 3.1 Pro has the strongest performance, generating a calibrated question in 44.9\% of sessions. DeepSeek V4 Pro and Qwen3.5-27B perform comparably, achieving 37.2\% and 37.4\% respectively, followed by Qwen3.5-4B at 27.2\%. Finally, Gemini 3.1 Flash Lite achieves the worst performance with a 25.2\% calibration rate. Figure \ref{fig:perf} also breaks down outcomes into questions that are too easy, too hard, missing, correctly calibrated, or rejected due to cross-check disagreement. As calibration performance worsens, the proportion of questions that are too easy increases consistently, suggesting that the primary failure mode of weaker models is generating insufficiently difficult questions. These results highlight two takeaways. First, more capable models perform better, suggesting that calibration correlates with broader model quality. Second, even state-of-the-art models fall well below perfect performance, demonstrating that Ask-E has substantial headroom to measure future improvements. Generation configurations are given in Appendix~\ref{sec:more_bench}.

\paragraph{Per Target Calibration.}
The 20 possible upper and lower bounds allow for 190 unique target difficulty ranges. Each asker model is evaluated 10 times on each range. Figure \ref{fig:heatmap} shows the calibration rate of Gemini 3.1 Pro for each pair, with boundary solvers ordered by approximate model size, increasing upward and leftward. Unsurprisingly, calibration is highest in the lower-left region of the heatmap, where one boundary solver is small and the other large, creating wide target ranges that are easier to hit. Performance is moderate in the lower-right region, where both models are small. These ranges are narrow but correspond to easier questions that may not require as much skill to generate. Calibration is lowest in the upper region, where both boundary solvers are highly capable. These pairs define narrow gaps that also demand generating very challenging questions.
The too-easy and too-hard heatmaps mirror this pattern. The asker is more likely to undershoot (too easy) when both boundary solvers are large, and to overshoot (too hard) when both are small.

\begin{figure*}[t]
  \centering
  \includegraphics[width=\textwidth]{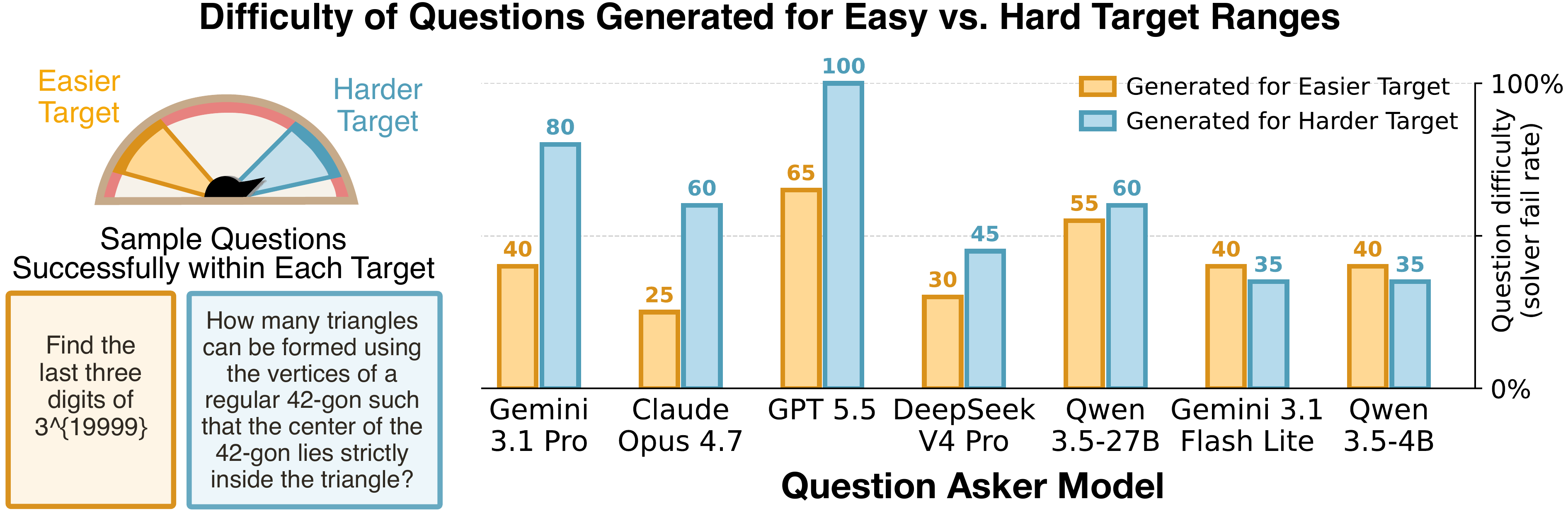}
   \caption{\textbf{Difficulty of questions generated for easy versus hard target ranges.} Each asker model generates questions targeting an easier range (Qwen3-1.7B to Gemma-3-4b-it) and a harder range (Gemini-3-Flash to Gemini-2.5-Flash). An independent solver model (Qwen3.5-2B) attempts each set; its fail rate serves as a proxy for question difficulty. Sample questions successfully calibrated within each target are shown on the left.}
   \label{fig:diff}
\end{figure*}

\paragraph{Measuring Question Difficulty.} Beyond calibration rate, we verify that models actually adjust question difficulty to the target rather than relying on a degenerate strategy, such as generating questions at a fixed difficulty that happens to fall within many ranges. To test this, we select two target ranges: an easy one (Qwen3-1.7B \citep{yang2025qwen3} vs.\ Gemma-3-4b-it \citep{gemma_2025}) and a hard one (Gemini-3-Flash \citep{gemini3flash2025} vs.\ Gemini-2.5-Flash \citep{comanici2025gemini}). Each asker generates questions for both ranges, and an independent solver (Qwen3.5-2B) attempts each set. Its failure rate serves as a proxy for question difficulty. As shown in Figure \ref{fig:diff}, frontier asker models produce substantially harder questions for the harder target, indicating genuine calibration, while less capable models show little or no difference between the two sets.

\paragraph{Extending the Benchmark.} We do not view the current benchmark formulation as fixed. As stronger models become available, the boundary set can be updated with more capable models, raising the difficulty ceiling without requiring any changes to the environment itself. Similarly, practical constraints on conversation length can be relaxed as technology advances, allowing for richer interactions and more nuanced calibration. We implement Ask-E so that boundary sets and dialogue specifications can be easily updated through configuration files, facilitating the evolution of the environment alongside the models it evaluates.

%% file: sections/training.tex
\section{RL Training on Ask-E}
In addition to a benchmark, Ask-E can be used as a training environment. The model is trained exclusively on question generation. It never receives feedback on the correctness of its own answers, and its reward is derived entirely from the answer distribution of solvers at or below its own ability. As we show in Section \ref{sec:results}, the skills developed through asking alone transfer to downstream question answering, improving the model's own performance on math benchmarks.

\subsection{Training Implementation}

\paragraph{Model and stack.}
We train Qwen3.5-4B (Instruct, no-thinking) as the asker policy with the \texttt{prime-rl}~\citep{primeintellect2025prime-rl} training library, which is natively compatible with the \texttt{verifiers}~\citep{brown_verifiers_2025} environment used to define our question-generation rollout. A single training run uses 11 H200 GPUs, partitioned across three roles: 4 GPUs serve the asker's rollout copy (inference for in-flight rollouts), 4 GPUs run the trainer copy of the asker together with optimizer state, and 3 GPUs host the small boundary-solver inference endpoints.

\begin{figure*}[t]
  \centering
  \includegraphics[width=\textwidth]{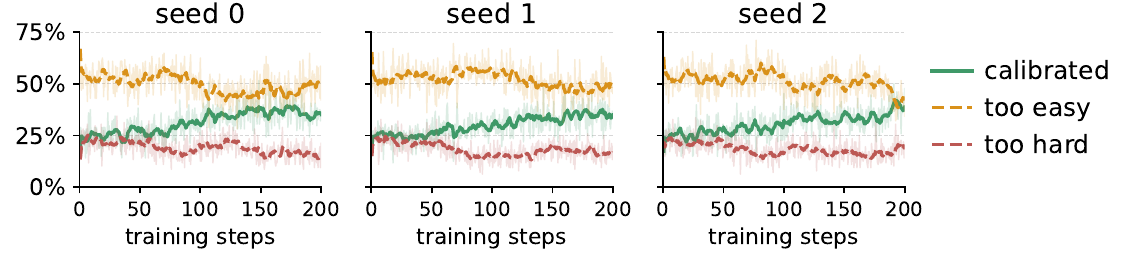}
   \caption{\textbf{Training curves across three seeds.} Calibration rate, too easy rate, and too hard rate over 200 training steps. All show improvement in calibration rate.}
   \label{fig:train}
\end{figure*}

\paragraph{Boundary set: weak only.}
To ensure that improvements come from the task itself rather than from learning from more capable models, the training boundary set contains only models that perform at or below the Qwen3.5-4B asker on math benchmarks (the 12-model Small set in Appendix~\ref{sec:sets}, 1.5B--8B open-weight checkpoints). The asker therefore never sees a stronger model's reasoning, ruling out implicit distillation as the source of any gains. Additionally, all tokens from the boundary solvers are masked during training; only the asker's own assistant-role tokens contribute to the loss, so no boundary-solver logits ever propagate back into the policy. Also note that the boundary solver weights remain frozen through training.

\paragraph{Self-judge cross-check.}
Whereas the benchmark uses a frontier model (Gemini 3.1 Pro) as the cross-check solver, during training the asker's own current weights serve as the cross-check solver. This ensures no information from a stronger model can leak into the policy through the reward channel. The cross-check occurs only after a rollout is complete, purely to compute the reward. Its output is never shown to the asker and never enters the interaction, so no cross-check tokens can influence generation or training beyond the scalar reward. Training also uses lighter-weight grading than the benchmark, a single cross-check sample per rollout rather than a consensus of three, and no LLM equivalence check, keeping the reward fast to compute.

\paragraph{Reward shape.} Per-rollout reward is the sum of three components. The primary signal is a calibration reward on the final question's outcome: $+1.0$ if exactly one boundary solver is correct (calibrated), $+0.2$ if both are correct (too easy), and $-0.2$ if neither is (too hard). We find rewarding too easy questions more than too hard questions is helpful, possibly because unclear questions tend to result in a too hard label. A format penalty of $-0.05$ is applied for each turn that does not contain the required tags. Finally, an embedding-diversity bonus encourages varied questions, detailed in Appendix~\ref{sec:more_training}.

\paragraph{Additional Training Configurations.}
We optimize the policy with CISPO~\citep{chen2025minimax}, using a batch size of 256 rollouts per optimizer step with group size $G=16$, covering 16 distinct boundary-pair examples per step. Training rollouts are shorter than the benchmark configuration for efficiency: 3 probing rounds plus 1 final question, with reduced token budgets. Training runs for 200 optimizer steps across 3 seeds. We report results aggregated across seeds. Full hyperparameters and optimizer details are in Appendix~\ref{sec:more_training}.

\subsection{Training Results}
\label{sec:results}

Over 200 steps, calibration rate improves consistently across all three seeds (Figure~\ref{fig:train}). More importantly, we find that the skills learned during training transfer to downstream tasks. To measure this, we evaluate the trained Ask-E model on four challenging math benchmarks, comparing performance to the Qwen3.5-4B model before training. Both models use the recommended Qwen3.5-4B sampling parameters for reasoning tasks in instruct mode. Answers are scored with rule-based equivalence matching, except on IMO-AnswerBench, where answers are graded with an LLM judge (additional details in Appendix~\ref{sec:more_math}). As shown in Table~\ref{tab:math-results-compact}, Ask-E training improves pass@8 performance on all benchmarks with one tie, and avg@8 performance on 3 out of 4 benchmarks. Notably, this improvement occurs without any new math data, any interaction with stronger models, or any feedback on the correctness of the trained model's answers.

\begin{table}[h]
\centering
\small
\setlength{\tabcolsep}{8pt}
\renewcommand{\arraystretch}{1.05}
\begin{tabular}{l l c c c}
\toprule
\textbf{Dataset} & \textbf{Metric} & \textbf{$n$} & \textbf{Qwen3.5-4B } & \textbf{with Ask-E} \\
\midrule
\multirow{2}{*}{AIME 2022--2024 \citep{aimo_validation_aime}}      & pass@8 & 90  & 87.78\% & \textbf{90.00\%} \\
                & avg@8  & 90  & 68.06\% & 67.36\%          \\
\midrule
\multirow{2}{*}{AIME 2025 \citep{opencompass_aime2025}}               & pass@8 & 30  & 83.33\% & 83.33\%          \\
          & avg@8  & 30  & 54.17\% & \textbf{57.22\%} \\
\midrule
\multirow{2}{*}{HMMT Feb 2025 \citep{dekoninck2026matharena}}               & pass@8 & 30  & 70.00\% & \textbf{72.22\%} \\
              & avg@8  & 30  & 45.00\% & \textbf{45.97\%} \\
\midrule
\multirow{2}{*}{IMO-AnswerBench \cite{luong2025towards}} & pass@8 & 400 & 70.25\% & \textbf{71.25\%} \\
                & avg@8  & 400 & 43.91\% & \textbf{44.90\%} \\
\bottomrule
\end{tabular}
\vspace{0.2cm}
\caption{Math benchmark performance. Qwen3.5-4B accuracy before Ask-E training vs. accuracy after Ask-E RL training. Ask-E values are averaged across
the 3 training seeds (full breakdown in Appendix~\ref{sec:more_table}). \textbf{Bold} marks improvements over the baseline. $n$ indicates benchmark size.}
\label{tab:math-results-compact}
\end{table}

%% file: sections/conclusion.tex
\section{Conclusion}
 
We have introduced Ask-E, an environment that benchmarks and trains models on calibrated question generation rather than question answering. Our results demonstrate that this task correlates with general model ability, remains far from saturated, and produces transferable skills through training. As models continue to improve, matching and surpassing human ability, we believe the field will need to move beyond standard supervised evaluation and training regimes, and we hope Ask-E represents a step in that direction.

%% file: supp/limits.tex
\section{Limitations and Broader Impact}
\label{sec:limits}

\paragraph{Limitations.} We identify two primary limitations of this work. First, there exist potential degenerate strategies that could satisfy the calibration criterion without requiring genuine mathematical skill. For example, an asker could pose a question with a random or ambiguous answer, such as asking models to flip a coin, which would produce answer variance without targeting any meaningful capability gap. The standard grading mitigates this directly: a session is only gradable when three independent cross-check samples agree, so questions without a single stable answer are rejected as bad cross-check rather than scored. We have not observed degenerate behavior in practice, and under 1\% of sessions are rejected by this filter, but subtler degenerate strategies remain possible in principle. Second, Ask-E is best suited for developing and measuring skills that a model already possesses to some degree. A model must be capable of generating reasonable questions to begin with in order to benefit from training in this environment, limiting its applicability for teaching entirely new skills from scratch.

\paragraph{Broader Impacts.} Our work is primarily foundational research into model evaluation and training. On the positive side, Ask-E offers a scalable approach to benchmarking and improving models that reduces dependence on expensive human-generated datasets. On the negative side, models that improve at generating calibrated questions could in principle be used to generate targeted evaluation or training data in ways that amplify existing biases in the boundary solvers. However, we view this risk as minimal given the current scope of the work, which is limited to math question generation.

%% file: supp/prompts.tex
\section{Environment Prompts}
\label{sec:prompts}

The environment runs a multi-turn dialogue between three roles:

\begin{itemize}
  \item \textbf{Asker} — generates probing
        questions and a final differentiating question. Receives a system
        prompt once, then a structured user message every turn.
  \item \textbf{Two boundary solvers} — answer each generated question.
        They share a single system prompt, see only the current question
        (no memory across turns), and reply with a boxed answer plus a
        brief summary that is fed back to the asker.
  \item \textbf{cross-check solver} — independently solves the final
        question to provide ground truth for the reward function. It uses
        a stripped-down ``grade'' system prompt: no \texttt{\#Summary\#}
        is needed because only the boxed answer is consumed.
\end{itemize}

Each rollout has $N$ probing turns followed by one final turn. The
asker's reply on every turn is parsed for three sections:
\texttt{\#Reasoning\#} (private), \texttt{\#Draft\#} (private), and
\texttt{\#Question\#} (forwarded verbatim to both boundary solvers).

\newtcblisting{promptbox}[2][]{%
  listing only,
  breakable,
  enhanced,
  colback=gray!4,
  colframe=gray!50,
  boxrule=0.5pt,
  arc=2pt,
  left=6pt, right=6pt, top=4pt, bottom=4pt,
  fonttitle=\bfseries,
  title={#2},
  listing options={%
    basicstyle=\small\ttfamily,
    breaklines=true,
    breakatwhitespace=false,
    columns=fullflexible,
    keepspaces=true,
    showstringspaces=false,
  },
  #1
}

\subsection*{Asker prompts}

\paragraph{System prompt (set once, at the start of every rollout).}
The asker is told the task, the constraints on its questions, and
the three-section response format it must emit every turn.

\begin{promptbox}{Asker system prompt}
TASK: Find a math question that exactly one of two models can solve correctly.

CONSTRAINTS:
- Questions must have a single unambiguous answer expressible in \boxed{} format (integer, fraction, decimal, or short expression).
- Answering models see ONLY the current question. They have no memory of prior rounds.
- 'Calibrated' means exactly one model is correct - both right or both wrong does not count.

STRATEGY:
You have {num_probing} probing rounds before your final question. Use them to find the difficulty boundary where one model fails and the other succeeds.

RESPONSE FORMAT (every turn, probing and final):

#Reasoning#
Reason through the previous round (if any) and decide what to target next. This is your private scratchpad - NOT sent to the models. Each round, consider:
- What concepts or difficulty levels seem to separate the two models
- What kinds of questions could effectively target that gap

#Draft#
Propose a candidate question and work through its solution end-to-end. If the answer is ambiguous, has multiple valid values, or depends on unstated assumptions, draft another and try again. End with \boxed{answer}. This section is NOT sent to the answering models.

#Question#
Your final question, copied cleanly from the draft. Everything after the #Question# tag is sent verbatim to both models, so stop generating once you've finished writing the question itself - do not include solutions, reasoning, hints, or any other text after the question.

Output exactly one #Reasoning# section, one #Draft# section, and one #Question# section per turn, in that order.
\end{promptbox}

\paragraph{First-turn user prompt.}
On round 1 the asker has no prior boundary-solver responses to react
to, so it is just shown the structure once more and asked for a probing
question.

\begin{promptbox}{First-turn user prompt}
Use #Reasoning# to plan your approach: think about what concepts or difficulty levels might separate the two models and what kinds of questions could effectively target those gaps. Use #Draft# to propose a candidate question and work through its solution. Then commit to your final probing question under #Question#.

EXAMPLE STRUCTURE:

#Reasoning#
(your reasoning)

#Draft#
(your candidate question + worked-out solution)

#Question#
(your question)
\end{promptbox}

\paragraph{Probing-round feedback (rounds $2 \ldots N$).}
After each probing round the asker is given both boundary solvers'
post-\texttt{\#Summary\#} replies and asked for the next probing
question.

\begin{promptbox}{Probing feedback (rounds 2..N)}
Model 1 answered:
{boundary_a_answer}

Model 2 answered:
{boundary_b_answer}

Use #Reasoning# to process these results: assess who got it right, what that reveals about each model's capabilities, and consider what to target next. Use #Draft# to propose a candidate question and work through its solution. Then commit to probing question number {next_question_num} under #Question#.

EXAMPLE STRUCTURE:

#Reasoning#
(your reasoning)

#Draft#
(your candidate question + worked-out solution)

#Question#
(your question)
\end{promptbox}

\paragraph{Final-question prompt.}
On the last turn the asker is told to synthesize and emit its final
differentiating question.

\begin{promptbox}{Final-question prompt}
Model 1 answered:
{boundary_a_answer}

Model 2 answered:
{boundary_b_answer}

This was your last probing round. Use #Reasoning# to:
- Synthesize what you have learned about each model's strengths and weaknesses
- Identify the specific gap your final question should exploit
- Brainstorm several candidate final questions and work through their solutions
- Pick the candidate most likely to be solved by exactly one model
You may reuse a probing question if you believe it discriminates well. Use #Draft# to propose your final question and work through its solution. Then commit to your final question under #Question#.

EXAMPLE STRUCTURE:

#Reasoning#
(your reasoning)

#Draft#
(your candidate question + worked-out solution)

#Question#
(your question)
\end{promptbox}

\paragraph{Recovery prompts.}
If the previous response did not contain a parseable
\texttt{\#Question\#}, the asker gets a one-turn recovery prompt
that nudges it to produce one. There are two variants — for probing
rounds and for the final round — each with two sub-cases (truncated vs.
malformed).

\begin{promptbox}{No-question recovery (probing round)}
[truncated case]
No question was found in your previous response - you ran out of tokens before producing a #Question# tag, and thus the models had no question to answer. Keep your reasoning shorter this time.

[malformed case]
No question was found in your previous response - you did not use the #Question# tag, and thus the models had no question to answer. You must output your question under the #Question# tag exactly as shown below.

Ask probing question number {next_question_num} of {num_probing} under #Question#.

EXAMPLE STRUCTURE:

#Reasoning#
(your reasoning)

#Draft#
(your candidate question + worked-out solution)

#Question#
(your question)
\end{promptbox}

\begin{promptbox}{No-question recovery (final round)}
[truncated case]
No question was found in your previous response - you ran out of tokens before producing a #Question# tag. Keep your reasoning much shorter this time.

[malformed case]
No question was found in your previous response - you did not use the #Question# tag. You must output your question under the #Question# tag exactly as shown below.

This was your last probing round. Use #Reasoning# to synthesize what you have learned, output the worked-out answer to your final question under #Solution#, and present your final differentiating question under #Question#.

EXAMPLE STRUCTURE:

#Reasoning#
(your reasoning)

#Draft#
(your candidate question + worked-out solution)

#Question#
(your question)
\end{promptbox}

\subsection*{Boundary-solver prompts}

\paragraph{Probing system prompt.}
Used during the probing rounds. Boundary solvers output a brief
\texttt{\#Summary\#} that is forwarded to the asker so it can
diagnose each model's reasoning without sending raw chains of thought
back upstream.

\begin{promptbox}{Boundary system prompt (probing)}
Answer the given question. Work through the problem, keeping your reasoning concise. When you are done, output #Summary# followed by your final answer in \boxed{} and a summary (3-10 sentences) of how you arrived at it.
\end{promptbox}

\paragraph{Question suffix (probing).}
Appended to every \texttt{\#Question\#} forwarded to the boundary solvers
during probing rounds, to reinforce the format.

\begin{promptbox}{Boundary question suffix (probing)}
When you are done, output #Summary# followed by your final answer in \boxed{} and a summary (3-10 sentences) of your approach.
\end{promptbox}

\subsection*{cross-check solver prompts}

\paragraph{Grading system prompt.}
The cross-check solver only needs to emit a boxed final answer for the
reward function — no summary section. The same prompt is reused if a
boundary solver is invoked in pure ``grade'' mode (e.g.\ when self-judge
is enabled).

\begin{promptbox}{Grading system prompt}
Answer the given question. Think step-by-step but keep your reasoning concise. Output the final solution in \boxed{}.
\end{promptbox}

\begin{promptbox}{Grading question suffix}
Output the final solution in \boxed{}.
\end{promptbox}

%% file: supp/full_boundary_models.tex
\section{Boundary Solver Sets}
\label{sec:sets}

We work with two boundary-solver collections. The \textbf{Full set} is
used at evaluation time to benchmark asker models; it spans 20
models from 1.5B parameters up to a frontier closed-API class. The
\textbf{Small set} is the subset used during RL training: 12 open-weight
models in the 1.5B--8B range. Every Small-set model is also in the Full
set.

\begin{table}[H]
\centering
\small
\setlength{\tabcolsep}{6pt}
\renewcommand{\arraystretch}{1.05}
\begin{tabular}{l l r c c}
\toprule
\textbf{Model} & \textbf{Family}  & \textbf{Params} & \textbf{Full} & \textbf{Small} \\
\midrule
\href{https://huggingface.co/deepseek-ai/DeepSeek-R1-Distill-Qwen-1.5B}{deepseek-r1-1.5b} \citep{guo2025deepseek}   & DeepSeek  & 1.5B & \checkmark & \checkmark \\
\href{https://huggingface.co/Qwen/Qwen3-1.7B}{qwen3-1.7b} \citep{yang2025qwen3}         & Qwen      & 1.7B & \checkmark & \checkmark \\
\href{https://huggingface.co/Qwen/Qwen3.5-2B}{qwen3.5-2b}  \citep{qwen3.5}        & Qwen     & 2.0B & \checkmark & \checkmark \\
\href{https://huggingface.co/meta-llama/Llama-3.2-3B}{llama-3.2-3b} \citep{grattafiori2024llama}      & Llama    & 3.0B & \checkmark & \checkmark \\
\href{https://huggingface.co/mistralai/Ministral-3-3B-Instruct-2512}{ministral-3-3b} \citep{liu2026ministral}      & Mistral   & 3.0B & \checkmark & \checkmark \\
\href{https://huggingface.co/microsoft/Phi-4-mini-instruct}{phi-4-mini}  \citep{abouelenin2025phi}       & Phi       & 3.84B & \checkmark & \checkmark \\
\href{https://huggingface.co/Qwen/Qwen3.5-4B}{qwen3.5-4b}  \citep{qwen3.5}        & Qwen     & 4.0B & \checkmark & \checkmark \\
\href{https://huggingface.co/Qwen/Qwen3-4B}{qwen3-4b} \citep{yang2025qwen3}            & Qwen    & 4.0B & \checkmark & \checkmark \\
\href{https://huggingface.co/google/gemma-3-4b-it}{gemma-3-4b-it} \citep{gemma_2025}     & Gemma    & 4.0B & \checkmark & \checkmark \\
\href{https://huggingface.co/deepseek-ai/DeepSeek-R1-Distill-Qwen-7B}{deepseek-r1-7b} \citep{guo2025deepseek}     & DeepSeek  & 7.0B & \checkmark & \checkmark \\
\href{https://huggingface.co/Qwen/Qwen3-8B}{qwen3-8b} \citep{yang2025qwen3}             & Qwen      & 8.0B & \checkmark & \checkmark \\
\href{https://huggingface.co/mistralai/Ministral-3-8B-Instruct-2512}{ministral-3-8b} \citep{liu2026ministral}      & Mistral  & 8.0B & \checkmark & \checkmark \\
\midrule
\href{https://huggingface.co/Qwen/Qwen3.5-9B}{qwen3.5-9b} \citep{qwen3.5}         & Qwen    & 9.0B  & \checkmark & \\
\href{https://huggingface.co/google/gemma-3-12b-it}{gemma-3-12b-it} \citep{gemma_2025}    & Gemma   & 12.0B & \checkmark & \\
\href{https://huggingface.co/Qwen/Qwen3-14B}{qwen3-14b} \citep{yang2025qwen3}            & Qwen     & 14.0B & \checkmark & \\
\href{https://huggingface.co/openai/gpt-oss-20b}{gpt-oss-20b}  \citep{openai2025gptoss120bgptoss20bmodel}      & GPT-OSS & 20.0B & \checkmark & \\
\href{https://huggingface.co/Qwen/Qwen3.5-27B}{qwen3.5-27b} \citep{qwen3.5}      & Qwen   & 27.0B & \checkmark & \\
\midrule
\href{https://docs.cloud.google.com/vertex-ai/generative-ai/docs/models/gemini/2-5-flash-lite}{gemini-2.5-flash-lite} \citep{comanici2025gemini} & Gemini & --- & \checkmark & \\
\href{https://docs.cloud.google.com/vertex-ai/generative-ai/docs/models/gemini/2-5-flash}{gemini-2.5-flash} \citep{comanici2025gemini}      & Gemini  & --- & \checkmark & \\
\href{https://docs.cloud.google.com/vertex-ai/generative-ai/docs/models/gemini/3-flash}{gemini-3-flash} \citep{gemini3flash2025}       & Gemini & --- & \checkmark & \\
\bottomrule
\end{tabular}
\vspace{0.5cm}
\caption{Boundary-solver sets. The \textbf{Full set} (20 models) is used
for benchmarking; the \textbf{Small set} (12 models) is the training
subset. Parameter counts are reported in billions; closed-API Gemini
Flash models do not have published sizes. Rows are ordered by
parameter count, with the Small set on top, the larger open-weight
models in the middle, and closed-API models at the bottom.}
\label{tab:boundary-sets}
\end{table}

\begin{table}[H]
\centering
\small
\setlength{\tabcolsep}{6pt}
\renewcommand{\arraystretch}{1.05}
\begin{tabular}{l l l}
\toprule
\textbf{Model} & \textbf{Family} & \textbf{License} \\
\midrule
deepseek-r1-1.5b      & DeepSeek & MIT + Apache~2.0 \\
qwen3-1.7b            & Qwen     & Apache~2.0 \\
qwen3.5-2b            & Qwen     & Apache~2.0 \\
llama-3.2-3b          & Llama    & Llama~3.2 Community \\
ministral-3-3b        & Mistral  & Apache~2.0 \\
phi-4-mini            & Phi      & MIT \\
qwen3.5-4b            & Qwen     & Apache~2.0 \\
qwen3-4b              & Qwen     & Apache~2.0 \\
gemma-3-4b-it         & Gemma    & Gemma Terms of Use \\
deepseek-r1-7b        & DeepSeek & MIT + Apache~2.0 \\
qwen3-8b              & Qwen     & Apache~2.0 \\
ministral-3-8b        & Mistral  & Apache~2.0 \\
\midrule
qwen3.5-9b            & Qwen     & Apache~2.0 \\
gemma-3-12b-it        & Gemma    & Gemma Terms of Use \\
qwen3-14b             & Qwen     & Apache~2.0 \\
gpt-oss-20b           & GPT-OSS  & Apache~2.0 \\
qwen3.5-27b           & Qwen     & Apache~2.0 \\
\midrule
gemini-2.5-flash-lite & Gemini   & Proprietary (API ToS) \\
gemini-2.5-flash      & Gemini   & Proprietary (API ToS) \\
gemini-3-flash        & Gemini   & Proprietary (API ToS) \\
\bottomrule
\end{tabular}
\vspace{0.5cm}
\caption{Licenses for each model in the boundary-solver sets.}
\label{tab:model-licenses}
\end{table}

%% file: supp/rollout_ex.tex
\section{Example Rollout}
\label{sec:rollout}

This section shows one full rollout from the Gemini 3.1 Pro benchmark
on the boundary pair \textbf{mistral-3-3b} vs.\ \textbf{qwen3\_5-27b}
(sample 2 of 10). For each turn we show the asker's complete
output (its \texttt{\#Reasoning\#}, \texttt{\#Draft\#}, and
\texttt{\#Question\#} sections), the question forwarded verbatim to
both boundary solvers, and the \emph{summaries} returned by each
boundary solver. On the final turn we additionally show the
cross-check solver's full response, which the reward function uses
as ground truth.

A boundary summary is constructed from the boundary solver's full
response by taking everything after the last \texttt{\#Summary\#}
(or \texttt{\#Output\#}) tag, truncated to 2000 characters; the
\texttt{\textbackslash boxed\{\}} final answer is always preserved
even if it falls outside the truncation window or sits above the
\texttt{\#Summary\#} tag. If the model emits no
\texttt{\#Summary\#}/\texttt{\#Output\#} tag, only the
\texttt{\textbackslash boxed\{\}} answer is forwarded; if neither is
present a placeholder string is forwarded instead. The full chains
of thought are never sent back to the asker.


\newtcblisting{questionbox}[2][]{%
  listing only, breakable, enhanced,
  colback=blue!3, colframe=blue!40, boxrule=0.5pt, arc=2pt,
  left=6pt, right=6pt, top=4pt, bottom=4pt,
  fonttitle=\bfseries, title={#2},
  listing options={basicstyle=\small\ttfamily, breaklines=true,
                   breakatwhitespace=false, columns=fullflexible,
                   keepspaces=true, showstringspaces=false},
  #1
}
\newtcblisting{summarybox}[2][]{%
  listing only, breakable, enhanced,
  colback=orange!4, colframe=orange!50, boxrule=0.5pt, arc=2pt,
  left=6pt, right=6pt, top=4pt, bottom=4pt,
  fonttitle=\bfseries, title={#2},
  listing options={basicstyle=\small\ttfamily, breaklines=true,
                   breakatwhitespace=false, columns=fullflexible,
                   keepspaces=true, showstringspaces=false},
  #1
}
\newtcblisting{keybox}[2][]{%
  listing only, breakable, enhanced,
  colback=green!3, colframe=green!50!black!60, boxrule=0.5pt, arc=2pt,
  left=6pt, right=6pt, top=4pt, bottom=4pt,
  fonttitle=\bfseries, title={#2},
  listing options={basicstyle=\small\ttfamily, breaklines=true,
                   breakatwhitespace=false, columns=fullflexible,
                   keepspaces=true, showstringspaces=false},
  #1
}

\subsection*{Turn 1 (probing)}

\begin{promptbox}{Asker output}
#Reasoning#
I'll start with a standard modular exponentiation problem to gauge their capabilities in number theory. The problem requires understanding of Euler's totient theorem or finding patterns in powers modulo 100. It's a classic competition-style problem, so both models might get it right, but if one fails, it gives me a clear idea of their upper bounds on number theory tasks.

#Draft#
Find the remainder when $2023^{2023}$ is divided by 100.

Solution:
We need to find $2023^{2023} \pmod{100}$. First, $2023 \equiv 23 \pmod{100}$. Since $\gcd(23, 100) = 1$, by Euler's totient theorem with $\phi(100) = 40$ we get $23^{40} \equiv 1 \pmod{100}$, so $23^{2023} \equiv 23^{23} \pmod{100}$. Computing successive squares: $23^2 \equiv 29$, $23^4 \equiv 41$, $23^{16} \equiv 61$. Then $23^{16}\cdot 23^4 \equiv 61\cdot 41 = 2501 \equiv 1$, so $23^{23} \equiv 23^3 \equiv 67 \pmod{100}$.
\boxed{67}

#Question#
Find the remainder when $2023^{2023}$ is divided by $100$.
\end{promptbox}

\begin{questionbox}{Question forwarded to both boundary solvers}
Find the remainder when $2023^{2023}$ is divided by $100$.
\end{questionbox}

\begin{summarybox}{mistral-3-3b summary returned to asker}
To find the remainder when $2023^{2023}$ is divided by 100, we first reduced the base modulo 100 to simplify the expression to $23^{2023} \mod 100$. Leveraging Euler's theorem, we simplified the exponent modulo $\phi(100)=40$, reducing the problem to computing $23^{23} \mod 100$. By using the Chinese Remainder Theorem to break down 100 into 4 and 25, we effectively solved the problem modulo these factors separately. Computing $23^{23} \mod 4$ and $23^{23} \mod 25$ gave us congruences $3 \mod 4$ and $17 \mod 25$, respectively. Combining these results using the CRT yielded the final solution, 67.
\boxed{67}
\end{summarybox}

\begin{summarybox}{qwen3\_5-27b summary returned to asker}
To find $2023^{2023} \pmod{100}$, we first reduced the base to $23 \pmod{100}$. Using Euler's Totient Theorem, we determined that the powers of 23 repeat with a period of $\phi(100)=40$. We reduced the exponent 2023 modulo 40 to get 23, simplifying the problem to $23^{23} \pmod{100}$. By calculating powers of 23 (specifically finding that $23^{20} \equiv 1$), we reduced the calculation to $23^3 \pmod{100}$. Finally, computing $23^3 = 12167$ gave a remainder of 67.
\boxed{67}
\end{summarybox}

\subsection*{Turn 2 (probing)}

\begin{promptbox}{Asker output}
#Reasoning#
Both models successfully solved the modular exponentiation problem, showing that they understand the Chinese Remainder Theorem, Euler's totient theorem, and basic modular arithmetic. I will now step up to a sequence problem that requires recognizing a pattern or finding a linear recurrence from a non-linear one. Nonlinear recurrences of the form $a_{n+1} = (a_n^2 + c)/a_{n-1}$ often reduce to a simple linear recurrence. Testing this will evaluate their algebraic manipulation and pattern recognition skills.

#Draft#
A sequence is defined by $a_1 = 1$, $a_2 = 2$, and $a_{n+1} = (a_n^2 + 1)/a_{n-1}$ for $n \ge 2$. Find $a_{10}$.

Calculate: $a_3 = 5$, $a_4 = 13$, $a_5 = 34$. The terms 1,2,5,13,34 are odd-indexed Fibonacci numbers. Equivalently, the sequence satisfies $a_{n+1} = 3a_n - a_{n-1}$. Iterating: $a_6=89$, $a_7=233$, $a_8=610$, $a_9=1597$, $a_{10}=4181$.
\boxed{4181}

#Question#
A sequence is defined by $a_1 = 1$, $a_2 = 2$, and $a_{n+1} = (a_n^2 + 1)/a_{n-1}$ for $n \ge 2$. Find the value of $a_{10}$.
\end{promptbox}

\begin{questionbox}{Question forwarded to both boundary solvers}
A sequence is defined by $a_1 = 1$, $a_2 = 2$, and $a_{n+1} = (a_n^2 + 1)/a_{n-1}$ for $n \ge 2$. Find the value of $a_{10}$.
\end{questionbox}

\begin{summarybox}{mistral-3-3b summary returned to asker}
The sequence was computed step-by-step using the recurrence relation $a_{n+1} = (a_n^2 + 1)/a_{n-1}$ with initial terms $a_1 = 1$ and $a_2 = 2$. An earlier error was identified in $a_8$, which was corrected to 610. Correcting back through earlier terms resolved inconsistencies. Finally, $a_{10}$ was computed exactly as 4181 by careful division.
\boxed{4181}
\end{summarybox}

\begin{summarybox}{qwen3\_5-27b summary returned to asker}
First, I calculated the first few terms of the sequence using the given recurrence relation $a_{n+1} = (a_n^2 + 1)/a_{n-1}$ to find $a_3=5$, $a_4=13$, $a_5=34$, $a_6=89$. Next, I observed that the terms satisfy the linear recurrence relation $a_{n+1} = 3a_n - a_{n-1}$, which is much easier to iterate than the original non-linear formula. Using this linear pattern, I computed the subsequent terms: $a_7=233$, $a_8=610$, $a_9=1597$. Finally, I calculated $a_{10} = 3(1597) - 610 = 4181$. Alternatively, recognizing the sequence as every second Fibonacci number ($a_n = F_{2n-1}$), I identified that $a_{10} = F_{19}$, which confirms the result is 4181.
\boxed{4181}
\end{summarybox}

\subsection*{Turn 3 (probing)}

\begin{promptbox}{Asker output}
#Reasoning#
Both models handled the sequence problem correctly, recognizing the linear recurrence or carefully computing terms. To push them further, I'll test their combinatorics and dynamic programming capabilities with a string avoidance problem. I'll ask for the number of sequences of a certain length avoiding a specific substring. This can be modeled with a recurrence relation or a state machine. It is straightforward but error-prone if the recurrence is built incorrectly.

#Draft#
Find the number of sequences of length 8 formed by the letters A, B, and C such that the sequence does not contain the consecutive substring "AB".

Solution:
Let $S_n$ be the total number of valid sequences of length $n$, and let $a_n, b_n, c_n$ be the count ending in A, B, C respectively. Appending A or C is unrestricted; appending B is forbidden after A. This gives $S_{n+1} = 3S_n - S_{n-1}$ with $S_1=3$, $S_2=8$. Iterating: $S_3=21$, $S_4=55$, $S_5=144$, $S_6=377$, $S_7=987$, $S_8=2584$.
\boxed{2584}

#Question#
Find the number of sequences of length 8 formed by the letters A, B, and C such that the sequence does not contain the consecutive substring "AB".
\end{promptbox}

\begin{questionbox}{Question forwarded to both boundary solvers}
Find the number of sequences of length 8 formed by the letters A, B, and C such that the sequence does not contain the consecutive substring "AB".
\end{questionbox}

\begin{summarybox}{mistral-3-3b summary returned to asker}
\boxed{1393}
This problem was solved using dynamic programming and recurrence relations to count valid sequences of length 8 formed by letters A, B, and C that avoid the forbidden substring "AB." The approach involved defining three separate counts for sequences ending in A, B, or C, then updating these counts iteratively based on the restrictions --- specifically, ensuring that a sequence cannot end with "AB." By breaking the problem into smaller subproblems (sequences of length 1 through 8), we accumulated the counts step-by-step, leveraging the relationships between sequences ending in A, B, or C. The solution efficiently explores all valid configurations without repetition, ensuring correctness through a systematic recurrence. The totals were computed up to length 8, yielding the final answer of 1,393 valid sequences.
\end{summarybox}

\begin{summarybox}{qwen3\_5-27b summary returned to asker}
To solve this problem, we defined a recurrence relation based on the last character of the sequence. Let $a_n$ be the total number of valid sequences of length $n$. We observed that a sequence ending in 'A' or 'C' can be formed by appending that letter to any valid sequence of length $n-1$, while a sequence ending in 'B' can only be formed by appending 'B' to a sequence not ending in 'A'. This led to the recurrence $a_n = 3a_{n-1} - a_{n-2}$. Starting with base cases $a_1 = 3$ and $a_2 = 8$, we iteratively calculated the values up to $n=8$. The sequence of counts is 3, 8, 21, 55, 144, 377, 987, 2584.
\boxed{2584}
\end{summarybox}

\subsection*{Turn 4 (probing)}

\begin{promptbox}{Asker output}
#Reasoning#
Model 2 perfectly recognized the recurrence relation and correctly computed the values, while Model 1 failed, likely due to a flawed DP state setup. This tells me Model 2 is quite strong at standard competition combinatorics and recursive logic, whereas Model 1 struggles when it has to correctly abstract a state machine. I will now test a probability problem involving expected value and Markov chains/state transitions. A classic problem is the expected number of rolls to get consecutive specific outcomes.

#Draft#
A fair standard 6-sided die is rolled repeatedly. What is the expected number of rolls needed to obtain two consecutive 6s?

Solution: Let $E_0$ be the expected rolls from the start and $E_1$ from after rolling a 6. Then $E_0 = 1 + (5/6)E_0 + (1/6)E_1$ and $E_1 = 1 + (5/6)E_0 + (1/6)\cdot 0$. Solving gives $E_0 = 42$.
\boxed{42}

#Question#
A fair standard 6-sided die is rolled repeatedly. What is the expected number of rolls needed to obtain two consecutive 6s?
\end{promptbox}

\begin{questionbox}{Question forwarded to both boundary solvers}
A fair standard 6-sided die is rolled repeatedly. What is the expected number of rolls needed to obtain two consecutive 6s?
\end{questionbox}

\begin{summarybox}{mistral-3-3b summary returned to asker}
The problem involves calculating the expected number of die rolls needed to obtain two consecutive sixes. This was approached by modeling the process using Markov states to capture the history of consecutive sixes observed. We defined two states: one where no six or only one six has appeared (State A), and another where a six has just appeared (State B). We set up equations for the expected number of rolls from each state, considering the probabilities of transitioning between states upon each roll. By solving the system of equations, we found the expected number of rolls needed to reach two consecutive sixes starting from the initial state to be 42.
\boxed{42}
\end{summarybox}

\begin{summarybox}{qwen3\_5-27b summary returned to asker}
To solve this problem, we defined two states: $E$ representing the expected rolls needed from the start, and $E_1$ representing the expected rolls needed after rolling a single 6. We set up a system of linear equations based on the transition probabilities between these states. Specifically, rolling a non-6 keeps us in the current state (or resets us to the start), while rolling a 6 advances us toward the goal. Solving the system $E = 1 + (1/6)E_1 + (5/6)E$ and $E_1 = 1 + (5/6)E$ yielded $E = 42$.
\boxed{42}
\end{summarybox}

\subsection*{Turn 5 (final)}

\begin{promptbox}{Asker output}
#Reasoning#
Through the probing rounds, both models demonstrated strong capabilities in modular arithmetic, recurrence relations, and expected value with state machines. However, in Round 3, Model 1 failed on a combinatorics problem involving strings that avoid a specific consecutive substring. Model 1 seems to struggle with setting up the correct dynamic programming states or principle of inclusion-exclusion for counting sequences with forbidden substrings. Model 2, on the other hand, excels at recognizing the exact recurrence relation and calculating the answer cleanly.

To exploit this gap, I will give a similar combinatorics string-avoidance problem but with slightly different parameters to ensure it cannot guess from memory. A 4-letter alphabet sequence of length 6 avoiding "AB" allows for verification via Principle of Inclusion-Exclusion (PIE) or a recurrence relation. Model 2 will likely solve it using the recurrence $S_n = 4 S_{n-1} - S_{n-2}$, while Model 1 will likely set up the DP incorrectly again.

#Draft#
Find the number of sequences of length 6 formed by the letters A, B, C, and D such that the sequence does not contain the consecutive substring "AB".

Solution: With four letters and one forbidden adjacency the recurrence becomes $S_{n+1} = 4 S_n - S_{n-1}$, with $S_0=1$, $S_1=4$. Iterating: $S_2=15$, $S_3=56$, $S_4=209$, $S_5=780$, $S_6=2911$. Verified independently by inclusion-exclusion over the 5 possible "AB" positions: $4096 - 1280 + 96 - 1 = 2911$.
\boxed{2911}

#Question#
Find the number of sequences of length 6 formed by the letters A, B, C, and D such that the sequence does not contain the consecutive substring "AB".
\end{promptbox}

\begin{questionbox}{Question forwarded to both boundary solvers}
Find the number of sequences of length 6 formed by the letters A, B, C, and D such that the sequence does not contain the consecutive substring "AB".
\end{questionbox}

\begin{summarybox}{mistral-3-3b summary returned to asker}
\boxed{2397}
\end{summarybox}

\begin{summarybox}{qwen3\_5-27b summary returned to asker}
\boxed{2911}
\end{summarybox}

\begin{keybox}{cross-check solver response (used as ground truth)}
To find the number of valid sequences of length 6, we can set up a recurrence relation. Let $S_n$ be the number of valid sequences of length $n$ that do not contain the consecutive substring "AB".

A valid sequence of length $n$ can be formed by appending a letter to a valid sequence of length $n-1$:
- We can append 'A', 'C', or 'D' to any valid sequence of length $n-1$. This gives $3 S_{n-1}$ valid sequences.
- We can append 'B' to any valid sequence of length $n-1$ *unless* it ends in 'A'. The number of valid sequences of length $n-1$ that end in 'A' is exactly the number of valid sequences of length $n-2$ (since the last letter is fixed as 'A'). Therefore, we can append 'B' in $S_{n-1} - S_{n-2}$ ways.

Adding these together, the recurrence relation for the total number of valid sequences is:
$$S_n = 3 S_{n-1} + (S_{n-1} - S_{n-2}) = 4 S_{n-1} - S_{n-2}$$

We can find the base cases easily:
- For $n=0$: $S_0 = 1$ (the empty sequence)
- For $n=1$: $S_1 = 4$ (A, B, C, D)

Now we can use the recurrence relation to find $S_6$:
- $S_2 = 4(4) - 1 = 15$
- $S_3 = 4(15) - 4 = 60 - 4 = 56$
- $S_4 = 4(56) - 15 = 224 - 15 = 209$
- $S_5 = 4(209) - 56 = 836 - 56 = 780$
- $S_6 = 4(780) - 209 = 3120 - 209 = 2911$

\boxed{2911}
\end{keybox}

%% file: supp/training_deats.tex
\section{Additional Training Details}
\label{sec:more_training}

\paragraph{Loss and clipping.}
We optimize the policy with the CISPO importance-clipped surrogate \citep{chen2025minimax}, using asymmetric clip bounds $\varepsilon_{\text{low}}=0.75$ and $\varepsilon_{\text{high}}=3.0$. The KL coefficient is fixed at zero. The loss is normalized per-token across the batch.

\paragraph{Optimizer.}
We use AdamW \citep{loshchilov2017decoupled} with learning rate \texttt{1e-6}, weight decay $0.01$, and gradient norm clipped at $1.0$. The learning rate uses a linear warmup of 60 steps with no decay thereafter.

\paragraph{Mixed-precision setup.}
Optimizer state is held in float32 with CPU offloading; the LM head is kept un-tied and in float32 separately from the model weights. Attention uses FlashAttention-3, with activation checkpointing on every transformer block. AdamW betas are $(\beta_1, \beta_2) = (0.9, 0.999)$ with $\varepsilon = 10^{-8}$.

\paragraph{Asynchronous rollouts.}
The trainer and inference pool run asynchronously with at most one trainer update fired before any in-flight rollout's log-probabilities are re-scored against the current policy. Sequence lengths are capped at $24{,}576$ tokens on the trainer side and $32{,}768$ tokens on the inference side.

\paragraph{Sampling.}
Rollout sampling uses temperature $1.0$ with the Qwen3.5-4B Instruct chat template (thinking disabled). The boundary solvers are queried at temperature $0.7$, and the self-judge cross-check solver is queried at temperature $0.0$ on the final turn.

\paragraph{Token budgets.}
Training rollouts use 4 turns (3 probing rounds + 1 final question). The asker is capped at 4{,}000 tokens during probing and the boundary solvers at 3{,}000. All other sampling parameters are left at the inference backend's defaults.

\paragraph{Self-judge cross-check.}
The prime-rl inference pool that hosts the policy doubles as the cross-check model. On the final turn, the question is sent (with a stripped grading prompt) to that same vLLM session at $\texttt{temperature}=0.0$. Only this single cross-check sample is used: unlike the benchmark grading, training applies no multi-sample consensus and no LLM equivalence check, and answers are compared with the symbolic equivalence pipeline alone.

\paragraph{Embedding-diversity bonus.}
Following \cite{hong2024curiosity}, each generated question is encoded with a small sentence embedder (\texttt{bge-small-en-v1.5}), and the bonus is $\alpha \cdot (1 - \overline{\text{cos-sim}}_{k\text{-NN}})$, where $\overline{\text{cos-sim}}_{k\text{-NN}}$ is the mean cosine similarity to the $k=5$ nearest neighbors in a sliding global buffer of the most recent $5{,}000$ questions and $\alpha=1.0$. We gate the diversity bonus on the calibration reward: the diversity term is multiplied by the (non-negative-clipped) calibration reward, so diversity is only rewarded when the question is calibrated. This prevents the policy from gaming the bonus by emitting novel but uncalibrated nonsense.

\paragraph{Seeds.}
Across the three seeds we vary both the rollout-example ordering and the inference-time sampling seed, so each seed differs in the order in which boundary pairs are presented and in the specific sequences sampled at each rollout step.

%% file: supp/bench_deats.tex
\section{Additional Ask-E Benchmark Implementation Details}
\label{sec:more_bench}

\paragraph{Asker-model configurations.}
Unless stated otherwise, we run asker models at the following thinking effort: \texttt{low} for all Gemini models, \texttt{medium} for Claude and GPT models, and \texttt{high} for DeepSeek. For all API-based askers (Gemini, DeepSeek, Claude \citep{claude_opus_4_7_systemcard}, GPT \citep{gpt5_systemcard}) we leave every sampling parameter at the provider's default and only override the thinking level. Qwen askers are run in instruct (no-thinking) mode with the sampling hyperparameters Qwen recommends for reasoning tasks in instruct mode: $\texttt{temperature}=1.0$, $\texttt{top\_p}=0.95$, $\texttt{top\_k}=20$, $\texttt{min\_p}=0.0$, $\texttt{presence\_penalty}=1.5$, and $\texttt{repetition\_penalty}=1.0$.

\paragraph{Token budgets.}
Per-turn token budgets are 6{,}000 tokens for the asker during probing and 10{,}000 for its final response; boundary solvers receive 4{,}000 tokens during probing and 6{,}000 on the final question.

\paragraph{Tag-based parsing.}
The text forwarded between roles is constructed by tag-based parsing, not by sending the raw model outputs verbatim:
\begin{itemize}[leftmargin=*]
  \item \textbf{Asker~$\rightarrow$~Boundary solvers.} Only the content following the asker's \texttt{\#Question\#} tag is forwarded; everything in \texttt{\#Reasoning\#} and \texttt{\#Draft\#} stays private. The forwarded question is truncated to 200 words to handle cases where the asker keeps generating text after the question itself, which would otherwise eat into the boundary solvers' token budget.
  \item \textbf{Boundary solvers~$\rightarrow$~Asker.} The boundary solver's full response is reduced to a short summary before being shown to the asker. We take everything after the model's last \texttt{\#Summary\#} (or \texttt{\#Output\#}) tag, truncated to 2{,}000 characters, with the \texttt{\textbackslash boxed\{\}} answer always preserved (re-prepended if it would otherwise fall outside the truncation window or sits above the \texttt{\#Summary\#} tag). If the model emits no \texttt{\#Summary\#}/\texttt{\#Output\#} tag at all, only the \texttt{\textbackslash boxed\{\}} answer is forwarded; if the response contains neither, a placeholder string is forwarded indicating the model failed to produce a structured answer.
\end{itemize}

If the asker fails to emit a parseable \texttt{\#Question\#} on the final turn (e.g.\ it runs out of tokens before reaching the tag), no boundary call is made and the session is recorded as \emph{no question}.

\paragraph{Scoring details.}
We use Gemini~3.1~Pro (\texttt{gemini-3.1-pro-preview}, \texttt{thinking\_level=low}, $\texttt{temperature}=0.7$) as the cross-check solver. On the final turn, the asker's question is sent independently to the cross-check solver, which solves it three times with independent samples. The three cross-check answers must agree with each other for the session to be gradable; sessions where they disagree are labeled \emph{bad cross-check} (at most 1.0\% of sessions for any asker), which filters out questions without a stable, unambiguous answer. A boundary solver is counted as correct only if its answer matches all three cross-check answers.

Final answers are extracted from both boundary responses and the cross-check response with the same procedure: scan for \texttt{\textbackslash boxed\{\dots\}} occurrences and return the contents of the last one. If no \texttt{\textbackslash boxed} appears, return the last non-punctuation word of the response. Empirically this means responses that hit the max-token limit are usually marked incorrect, but we find this acceptable: models that are uncertain tend to repeat flawed reasoning rather than commit to a wrong final answer, so frequently running out of tokens on a given question is itself a strong signal that the model could not solve it.

Equivalence is checked with the \texttt{math-verify} library \citep{kydlicek2025mathverify}, preceded by a normalization pass that strips \texttt{\textbackslash boxed}, \texttt{\textbackslash text}/\texttt{\textbackslash mathrm} wrappers, math delimiters, \texttt{\textbackslash left}/\texttt{\textbackslash right} pairs, and LaTeX spacing commands. If symbolic verification fails we fall through to whitespace- and case-insensitive string equality, comma-separated set equality (for reordered solution sets), and text-only fuzzy matching for ``no solution'' synonyms. Finally, answer pairs that the symbolic pipeline marks as non-matching are reviewed by an LLM equivalence checker (Gemini~3.1~Pro, \texttt{thinking\_level=low}), which promotes answers that are mathematically equivalent to the cross-check answer but differ in formatting (e.g., ``1'' vs.\ ``$x=1$'', or an unreduced fraction). A session is labeled \emph{calibrated} when exactly one of the two boundary solvers matches the cross-check answer, \emph{too\_easy} when both match, and \emph{too\_hard} when neither matches.

%% file: supp/additional_analyses.tex
\section{Additional Benchmark Analyses}
\label{sec:additional_analyses}

\subsection{The Role of the Cross-Check Solver}
\label{sec:cross_check_role}

In standard automated benchmarks, an answer produced by a grading model serves as an estimate of an independently existing ground truth, and any error in that estimate directly corrupts the benchmark label. The cross-check solver in Ask-E plays a different role. The target of a session is not a ground-truth value but a distribution of model answers: a session succeeds when the cross-check answers agree with exactly one boundary solver and disagree with the other. The cross-check answer therefore acts not as a verifier of the asker's output but as one component of the answer distribution the asker must elicit.

Consider writing an exam question for a class while aiming for a specific difficulty. One approach is to solve the question yourself, observe that the answer is A, and check that some students answered A while the rest scattered. Another is to never determine the correct answer, and instead check whether part of the class converged on a single answer while the rest scattered. The second admits some noise, but eliciting that distribution requires essentially the same skill as the first. This is the sense in which Ask-E's reward is not correctness-based: nothing the asker outputs is ever checked for correctness, and the models' joint answer behavior is itself the observable being measured. The consensus of three cross-check samples (Section~\ref{sec:benchmark}) reduces the noise in this distribution, and Appendix~\ref{sec:opus_check} shows the results remain stable even when the cross-check solver is replaced entirely.

\subsection{Robustness to the Choice of Cross-Check Solver}
\label{sec:opus_check}

\begin{figure*}[t!]
  \centering
  \includegraphics[width=\textwidth]{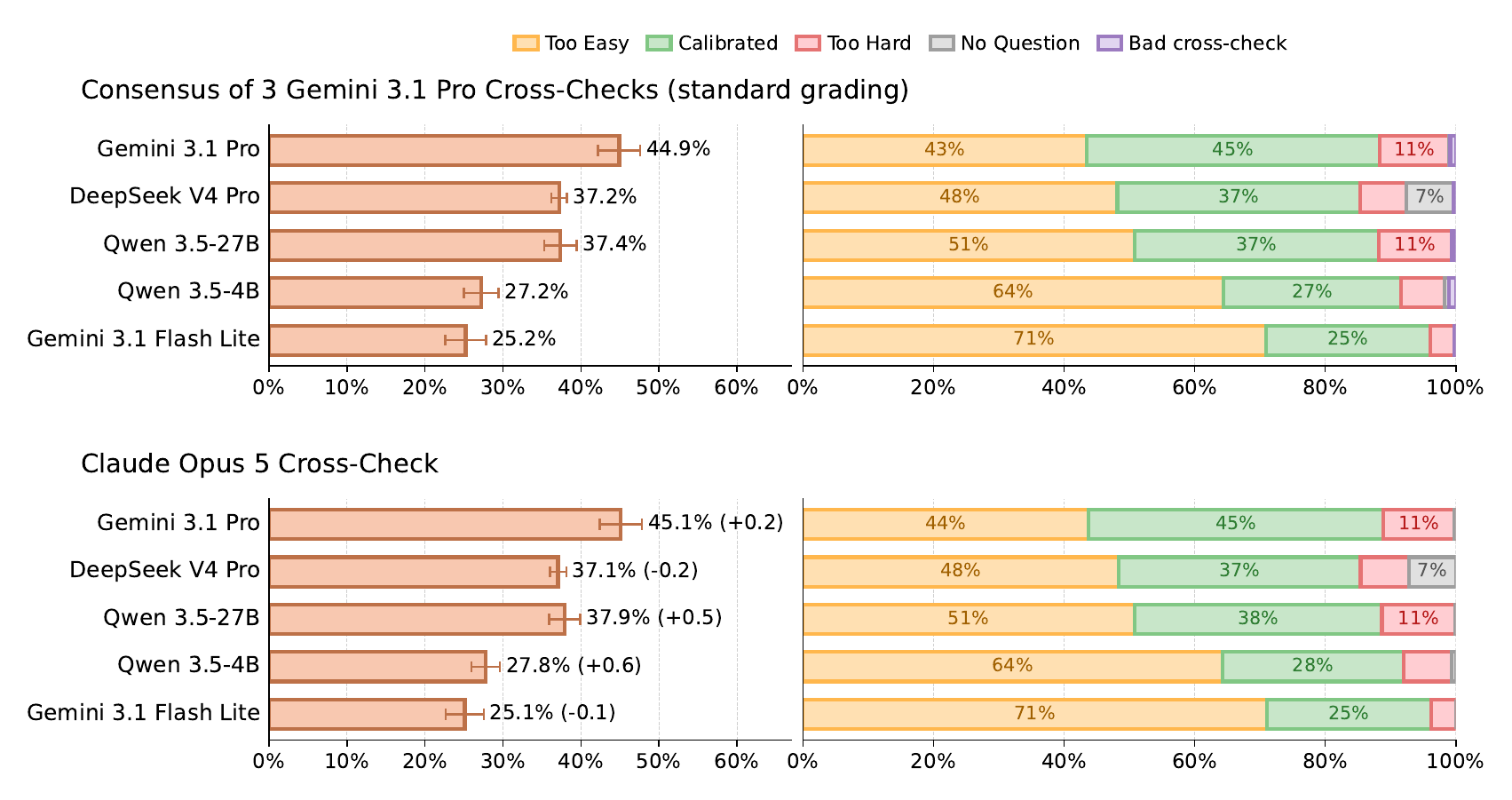}
   \caption{\textbf{Benchmark outcomes under two cross-check solvers.} (Top) The standard grading, a consensus of three independent Gemini 3.1 Pro cross-check samples. (Bottom) A single Claude Opus 5 cross-check, with each asker's calibrated rate annotated with its difference from the standard grading. Calibration rates shift by at most 0.6 percentage points and the model ranking is unchanged.}
   \label{fig:opus_check}
\end{figure*}

Because all grading in Ask-E is derived from model behavior, a natural question is whether benchmark results depend on the specific model used as the cross-check solver. To test this, we regrade every session of the main benchmark using Claude Opus 5 \citep{claude_opus_5_systemcard}, a model from a different family than every asker and boundary solver, as the cross-check solver in place of Gemini 3.1 Pro. Figure~\ref{fig:opus_check} compares outcomes under the standard grading (consensus of three Gemini 3.1 Pro cross-checks) against the Opus 5 grading, which uses the same symbolic and LLM equivalence checking. The calibration rate of every asker changes by at most 0.6 percentage points, and the ranking of askers is unchanged. The benchmark's measurements therefore reflect the interaction between the generated questions and the boundary solvers rather than any artifact of the grading model.

\subsection{No-Probing Baseline}
\label{sec:no_probe}

\begin{figure*}[t!]
  \centering
  \includegraphics[width=\textwidth]{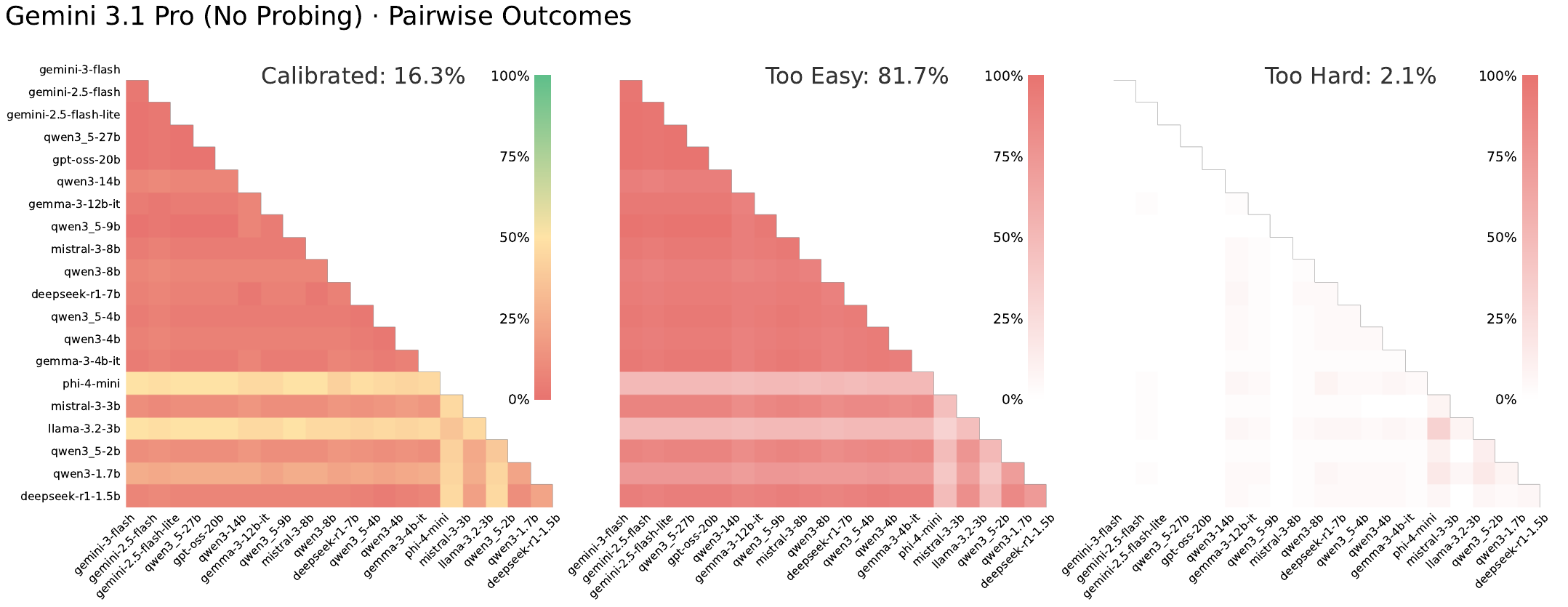}
   \caption{\textbf{Pairwise outcomes for the no-probing baseline.} Gemini 3.1 Pro generates 50 questions with the same task description but no probing rounds, and each question is graded against all 190 boundary pairs under the standard grading. The calibration rate falls from 44.9\% (with probing) to 16.3\%, with 81.7\% of outcomes too easy. The remaining calibrated outcomes concentrate in pairs containing the weakest boundary solvers.}
   \label{fig:no_probe}
\end{figure*}

To measure how much of benchmark performance comes from probing, we evaluate a baseline in which the asker receives the same task description but no probing rounds. Without probing, the asker receives no information about the specific boundary pair, so we have Gemini 3.1 Pro generate 50 questions unconditioned on any pair and evaluate each question against all 190 boundary pairs under the standard grading. The calibration rate drops from 44.9\% with probing to 16.3\% without, showing that the majority of benchmark performance comes from probing and pair-specific calibration rather than from generating generically difficult questions. Figure~\ref{fig:no_probe} shows the resulting pairwise outcomes: 81.7\% of outcomes are too easy, and the calibration that remains concentrates in pairs containing the weakest boundary solvers, where unconditioned questions happen to fall within wide capability gaps.

\subsection{Direction of Calibrated Outcomes}
\label{sec:weaker_wins}

The calibration criterion is symmetric: a session counts as calibrated whichever boundary solver answers correctly. This is deliberate. Requiring the stronger solver to be the correct one would presuppose a trusted external ranking of the two models, precisely the supervision Ask-E aims to avoid. Moreover, model ability is not a total order: a solver that is stronger overall may still fail on a specific question that a weaker solver handles, especially when the two are close in capability.

\begin{figure*}[t!]
  \centering
  \includegraphics[width=0.49\textwidth]{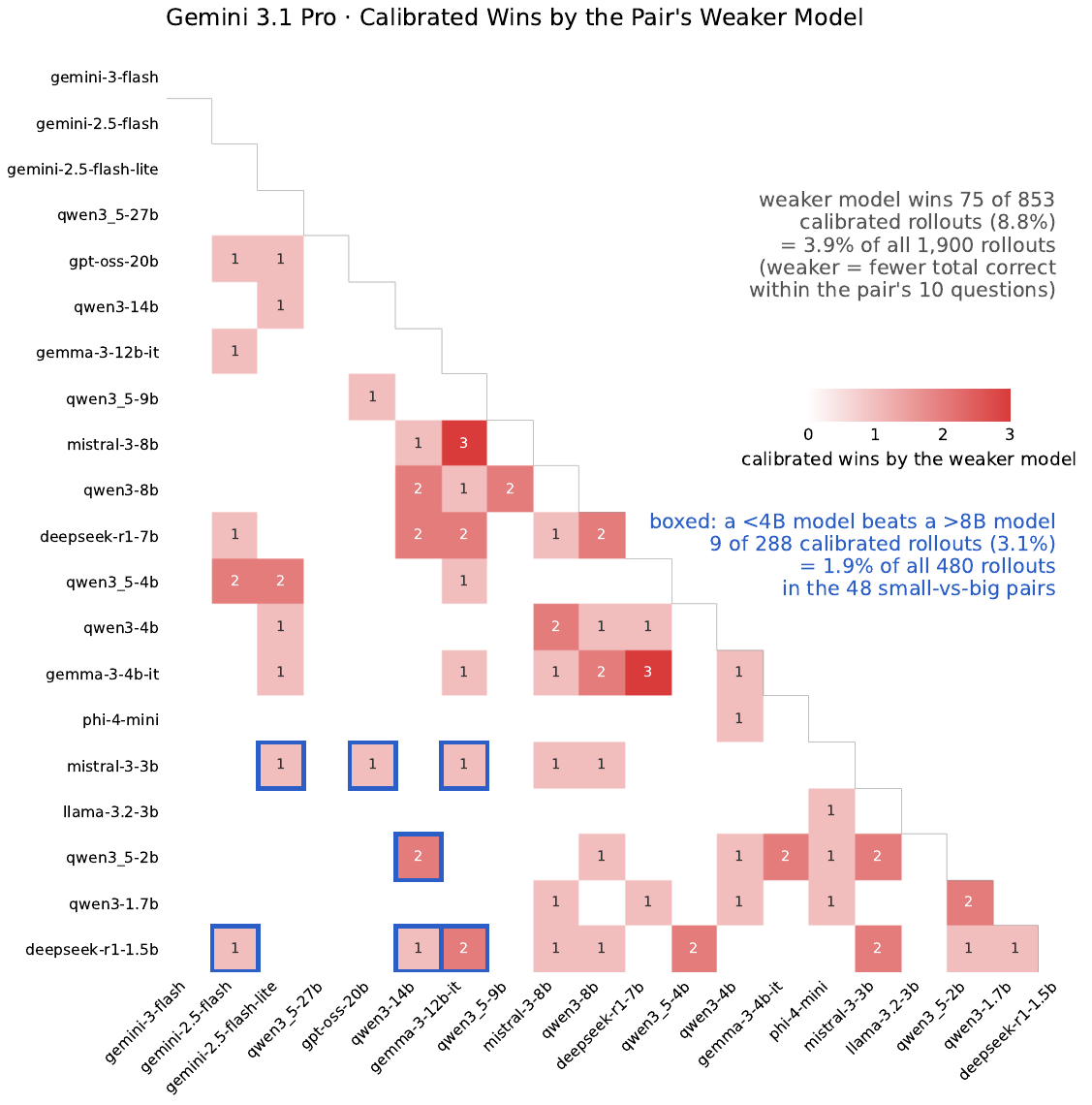}\hfill
  \includegraphics[width=0.49\textwidth]{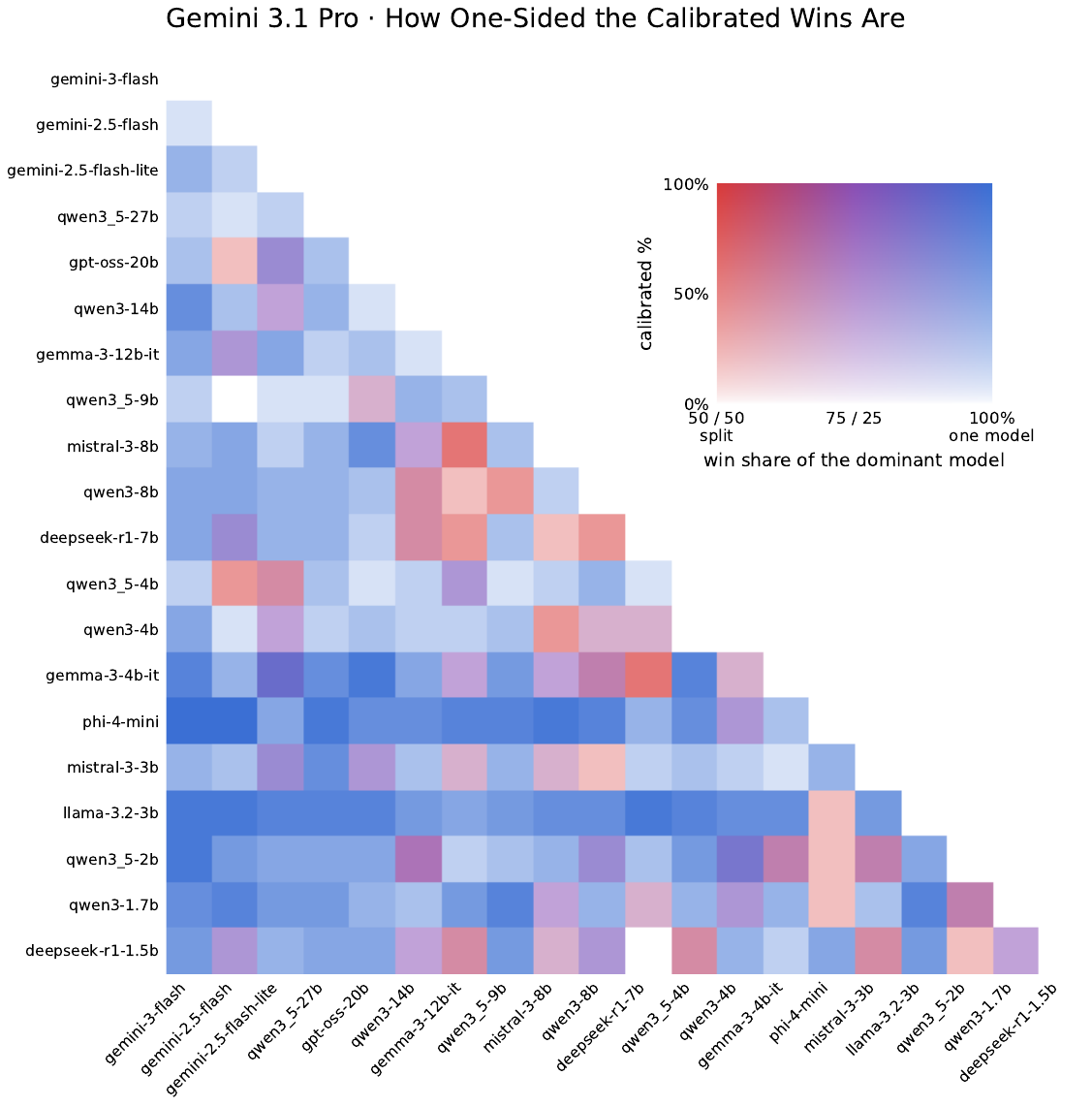}
   \caption{\textbf{Direction of calibrated outcomes.} (Left) Calibrated wins by the pair's weaker solver for the Gemini 3.1 Pro asker: each cell counts the calibrated sessions in which the empirically weaker solver of the pair (fewer total correct answers across the pair's ten questions) is the one that answers correctly, occurring in 75 of 853 calibrated sessions (8.8\%). Blue boxes mark clearly separated pairs (one solver under 4B parameters, the other over 8B), where it occurs in 9 of 288 calibrated sessions (3.1\%). (Right) How one-sided calibrated outcomes are: for each pair, color encodes the win share of the dominant solver among the pair's calibrated sessions (blue: one solver wins nearly all; red: an even split), with intensity scaled by how often the pair is calibrated at all. Most pairs are strongly one-sided, and even splits concentrate among pairs of similar capability.}
   \label{fig:weaker_wins}
\end{figure*}

Empirically, the split overwhelmingly falls in the expected direction. Figure~\ref{fig:weaker_wins} (left) measures, for the Gemini 3.1 Pro asker, how often the empirically weaker solver of a pair (fewer total correct answers across the pair's ten questions) is the one that answers correctly in a calibrated session. This occurs in 75 of 853 calibrated sessions (8.8\%). Restricting to clearly separated pairs, with one solver under 4B parameters and the other over 8B, it occurs in 9 of 288 calibrated sessions (3.1\%). Figure~\ref{fig:weaker_wins} (right) complements this by showing, for each pair, how close the calibrated outcomes are to one solver dominating versus an even split. Most pairs are strongly one-sided, indicating that the asker finds a consistent capability gap rather than coin-flip noise, and even splits concentrate among pairs of similar ability, exactly where a strict ordering is least meaningful.

%% file: supp/downstream_stuff.tex
\section{Additional Downstream Math Benchmark Implementation Details}
\label{sec:more_math}

We evaluate the trained asker (Qwen3.5-4B Instruct, no-thinking)
and the un-trained Qwen3.5-4B Instruct baseline on the same downstream
math benchmark suite, using identical sampling configurations.

\paragraph{Datasets.}
We evaluate on four public math reasoning benchmarks. All datasets
are loaded from their respective HuggingFace hubs:

\begin{table}[h]
\centering
\small
\setlength{\tabcolsep}{6pt}
\renewcommand{\arraystretch}{1.05}
\begin{tabular}{l c l}
\toprule
\textbf{Dataset} & \textbf{$n$} & \textbf{HuggingFace ID} \\
\midrule
AIME (2022--2024) & 90  & \href{https://huggingface.co/datasets/AI-MO/aimo-validation-aime}{\texttt{AI-MO/aimo-validation-aime}} \\
AIME 2025         & 30  & \href{https://huggingface.co/datasets/opencompass/AIME2025}{\texttt{opencompass/AIME2025}} (parts I + II) \\
HMMT 2025         & 30  & \href{https://huggingface.co/datasets/MathArena/hmmt_feb_2025}{\texttt{MathArena/hmmt\_feb\_2025}} \\
IMO AnswerBench   & 400 & \href{https://huggingface.co/datasets/Hwilner/imo-answerbench}{\texttt{Hwilner/imo-answerbench}} \\
\bottomrule
\end{tabular}
\caption{Downstream math benchmarks used to evaluate the trained
asker against the Qwen3.5-4B Instruct baseline. Counts are number
of problems; AIME 2022--2024 is the AIMO validation pool of 90
problems, AIME 2025 is the union of parts I and II, HMMT 2025 is the
February problem set, and IMO AnswerBench is the full release.}
\label{tab:math-eval-datasets}
\end{table}

\paragraph{Sampling.}
Both the baseline and trained models are evaluated with the same
sampling configuration to keep the comparison apples-to-apples.
Specifically, we use the Qwen recommended hyperparameters for
reasoning tasks in instruct mode (matching the asker sampling
described in Section~\ref{sec:benchmark}):
$\texttt{temperature}=1.0$, $\texttt{top\_p}=0.95$,
$\texttt{top\_k}=20$, $\texttt{min\_p}=0.0$,
$\texttt{presence\_penalty}=1.5$, and
$\texttt{repetition\_penalty}=1.0$, with thinking disabled (instruct
chat template, no \texttt{<think>} block),
$\texttt{max\_tokens}=32{,}000$, and $n=8$ samples per problem.

We do not use any system prompt and we do not provide few-shot
examples; the problem is sent as a single user message with a short
instruction appended that tells the model to reason step-by-step and
emit its final answer in \texttt{\textbackslash boxed\{\}}. The exact
instruction depends on whether the dataset has integer answers:

\begin{itemize}[leftmargin=*]
  \item \textbf{AIME and AIME 2025} (integer answers in $[0, 999]$):
        ``\textit{Please solve this problem step by step. The answer
        is an integer from 0 to 999. Put your final answer in
        \texttt{\textbackslash boxed\{\}}.}''
  \item \textbf{HMMT 2025 and IMO AnswerBench} (free-form answers):
        ``\textit{Please solve this problem step by step and put your
        final answer in \texttt{\textbackslash boxed\{\}}.}''
\end{itemize}

The instruction is appended to the raw problem text and the
combined string is the user-role content sent to the model's chat
template (with thinking disabled).

\paragraph{Metrics.}
For each dataset we report two metrics computed over the 8 samples
per problem:

\begin{itemize}[leftmargin=*]
  \item \textbf{pass@8}: 1 if at least one of the 8 samples for that
        problem produces the correct final answer, 0 otherwise.
        Averaged over problems.
  \item \textbf{avg@8}: per-problem fraction of correct samples,
        averaged over problems (equivalently, the per-sample
        accuracy averaged across the dataset).
\end{itemize}

\paragraph{Answer extraction.}
We extract the final answer from each model response with the same
procedure used elsewhere in the paper: we scan the response for
\texttt{\textbackslash boxed\{\dots\}} occurrences and return the
contents of the last one. If no \texttt{\textbackslash boxed} is
present, we fall back to the last non-punctuation word. Responses
that hit the 32{,}000-token cap without producing a
\texttt{\textbackslash boxed} answer are scored as incorrect (in
practice, models that have to be cut off rarely commit to a final
answer at all).

\paragraph{Equivalence checking.}
For AIME, AIME 2025, and HMMT 2025, we use rule-based equivalence
matching (the \texttt{math-verify}~\citep{kydlicek2025mathverify}
library combined with our LaTeX normalization pass) since their
answers are integers or short closed-form expressions. For IMO
AnswerBench, the ground-truth answers are frequently LaTeX
expressions whose surface form varies substantially across valid
restatements (e.g.\ \texttt{1+\textbackslash sqrt\{5\}/2} vs.\
\texttt{\textbackslash frac\{1+\textbackslash sqrt\{5\}\}\{2\}}), so
rule-based matching under-counts correct answers. We therefore
\textbf{re-grade IMO AnswerBench with an LLM judge}: each
(problem, ground-truth, candidate) triple is sent to
\texttt{gemini-3.1-flash-lite-preview} with a fixed yes/no prompt
asking whether the student answer is mathematically equivalent to
the ground truth, and the judge's verdict is taken as correctness.
The IMO AnswerBench numbers reported in Table~\ref{tab:math-results}
are the LLM-judge accuracies.

%% file: supp/full_table.tex
\section{Full Per-Seed Training Results}
\label{sec:more_table}

\begin{table}[h]
\centering
\small
\setlength{\tabcolsep}{6pt}
\renewcommand{\arraystretch}{1.05}
\begin{tabular}{l l c c c c c c}
\toprule
\textbf{Dataset} & \textbf{Metric} & \textbf{$n$} & \textbf{Baseline} & \textbf{Mean$_3$} & \textbf{seed 0} & \textbf{seed 1} & \textbf{seed 2} \\
\midrule
AIME (2022--2024)       & pass@8 & 90  & 87.78\% & \textbf{90.00\%} & \textbf{93.33\%} & 86.67\%          & \textbf{90.00\%} \\
AIME (2022--2024)       & avg@8  & 90  & 68.06\% & 67.36\%          & 67.08\%          & \textbf{68.19\%} & 66.81\%          \\
\midrule
AIME 2025               & pass@8 & 30  & 83.33\% & 83.33\%          & 83.33\%          & 83.33\%          & 83.33\%          \\
AIME 2025               & avg@8  & 30  & 54.17\% & \textbf{57.22\%} & \textbf{58.33\%} & \textbf{54.58\%} & \textbf{58.75\%} \\
\midrule
HMMT 2025               & pass@8 & 30  & 70.00\% & \textbf{72.22\%} & \textbf{73.33\%} & \textbf{73.33\%} & 70.00\%          \\
HMMT 2025               & avg@8  & 30  & 45.00\% & \textbf{45.97\%} & \textbf{45.83\%} & \textbf{45.83\%} & \textbf{46.25\%} \\
\midrule
IMO AnswerBench & pass@8 & 400 & 70.25\% & \textbf{71.25\%} & \textbf{72.00\%} & \textbf{71.50\%} & 70.25\%          \\
IMO AnswerBench & avg@8  & 400 & 43.91\% & \textbf{44.90\%} & \textbf{45.28\%} & \textbf{45.47\%} & \textbf{43.94\%} \\
\bottomrule
\end{tabular}
\caption{Math-benchmark accuracy of Qwen3.5-4B before Ask-E training vs.\ after Ask-E RL training. Mean$_3$ is the mean across the 3 training seeds; \textbf{bold} marks
values that beat the baseline. IMO AnswerBench is graded with the LLM
judge (numbers are pass@8 and avg@8 over 8 samples per problem).}
\label{tab:math-results}
\end{table}

\begin{figure*}[t!]
  \centering
  \includegraphics[width=\textwidth]{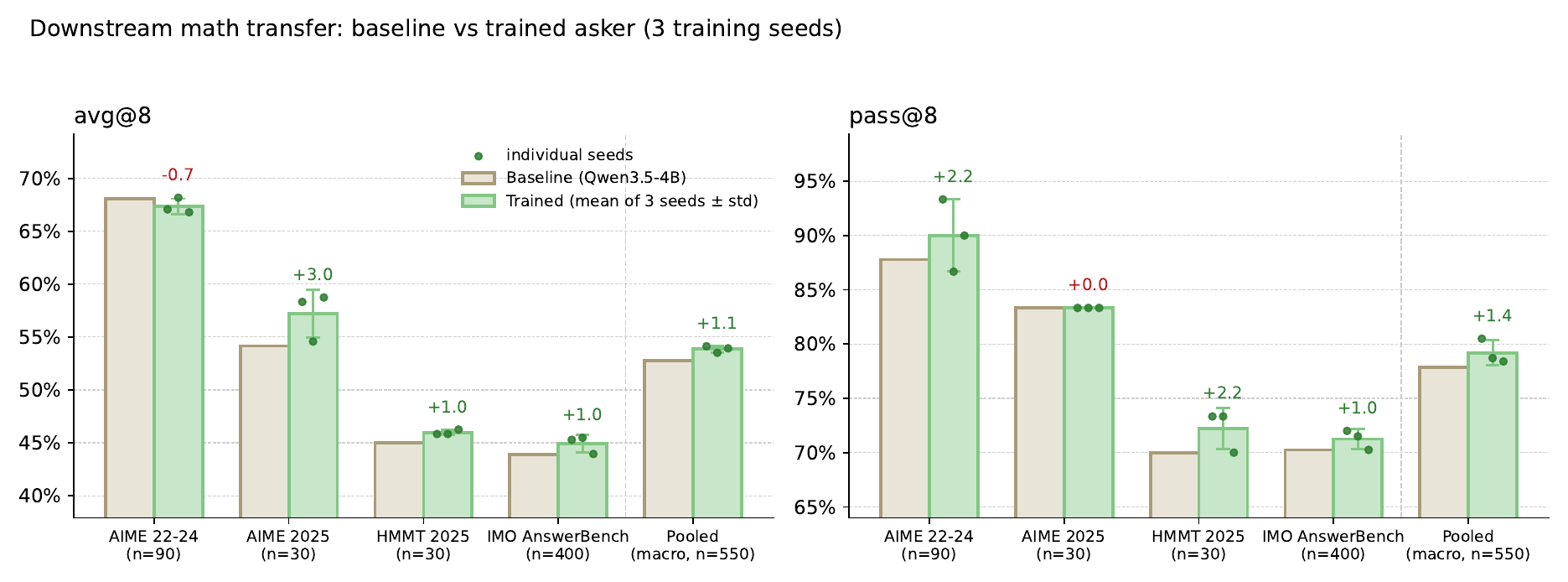}
   \caption{\textbf{Downstream math transfer per benchmark and pooled over the 550-problem suite.} Bars show the untrained Qwen3.5-4B baseline and the mean of the three trained seeds; whiskers show $\pm$ one standard deviation across seeds, and dots mark individual seeds. Annotations give the change from the baseline in percentage points.}
   \label{fig:seed_bars}
\end{figure*}

Figure~\ref{fig:seed_bars} summarizes transfer per benchmark and pooled across the full 550-problem suite, with each training seed shown individually and bars showing the mean $\pm$ one standard deviation across the three seeds. On the pooled suite, every seed improves over the baseline on both metrics, and for avg@8 the per-seed spread (0.6pp, with gains ranging from $+0.7$ to $+1.4$pp) is smaller than the mean gain of $+1.1$pp. The per-benchmark pattern follows benchmark difficulty. On AIME 2022--2024, the easiest benchmark in the suite (baseline avg@8 of 68\%), performance is roughly flat, with avg@8 changing by $-0.7$pp while pass@8 improves by $+2.2$pp. The harder benchmarks improve on avg@8, with AIME 2025 gaining $+3.0$pp, HMMT 2025 $+1.0$pp, and IMO AnswerBench $+1.0$pp. While these gains are modest, they are obtained with no new math data, no interaction with stronger models, and no correctness reward, and the consistent per-seed direction on the pooled suite indicates the transfer is real rather than seed noise.

